%% file: main.tex
\pdfoutput=1

\documentclass[11pt,dvipsnames]{article}

\usepackage[final]{acl}

\usepackage{times}
\usepackage{latexsym}
\usepackage[most]{tcolorbox}
\tcbuselibrary{listings, breakable}
\usepackage{subcaption}
\usepackage{xspace}
\usepackage{xurl}
\usepackage{hyperref}

\usepackage[T1]{fontenc}

\usepackage[utf8]{inputenc}

\usepackage{microtype}

\usepackage{inconsolata}

\usepackage{graphicx}
\usepackage{tabularx}
\usepackage{ragged2e}  
\usepackage{multirow}

\newcolumntype{Y}[1]{>{\RaggedRight\arraybackslash}p{#1}}

\usepackage{amsmath}
\usepackage{amsfonts}

\usepackage{booktabs}
\usepackage[textsize=tiny]{todonotes}

\def\ddefloop#1{\ifx\ddefloop#1\else\ddef{#1}\expandafter\ddefloop\fi}
\def\ddef#1{\expandafter\def\csname bb#1\endcsname{\ensuremath{\mathbb{#1}}}}
\ddefloop ABCDEFGHIJKLMNOPQRSTUVWXYZ\ddefloop

\def\ddef#1{\expandafter\def\csname c#1\endcsname{\ensuremath{\mathcal{#1}}}}
\ddefloop ABCDEFGHIJKLMNOPQRSTUVWXYZ\ddefloop

\def\ddef#1{\expandafter\def\csname v#1\endcsname{\ensuremath{\boldsymbol{#1}}}}
\ddefloop ABCDEFGHIJKLMNOPQRSTUVWXYZabcdefghijklmnopqrstuvwxyz\ddefloop

\def\ddef#1{\expandafter\def\csname v#1\endcsname{\ensuremath{\boldsymbol{\csname #1\endcsname}}}}
\ddefloop {alpha}{beta}{gamma}{delta}{epsilon}{varepsilon}{zeta}{eta}{theta}{vartheta}{iota}{kappa}{lambda}{mu}{nu}{xi}{pi}{varpi}{rho}{varrho}{sigma}{varsigma}{tau}{upsilon}{phi}{varphi}{chi}{psi}{omega}{Gamma}{Delta}{Theta}{Lambda}{Xi}{Pi}{Sigma}{varSigma}{Upsilon}{Phi}{Psi}{Omega}{ell}\ddefloop

\DeclareMathOperator*{\argtopk}{arg\,topK}

\def\1{\mathds{1}}

\newcommand{\ours}{\textsc{RI\textsuperscript{2}VER}\xspace}

\title{Inter-Passage Verification for Multi-evidence Multi-answer QA}

\author{
 \textbf{Bingsen Chen\textsuperscript{1,2}},
 \textbf{Shengjie Wang\textsuperscript{1,2}},
 \textbf{Xi Ye\textsuperscript{3}},
 \textbf{Chen Zhao\textsuperscript{1,2}}
\\
 \textsuperscript{1}NYU Shanghai,
 \textsuperscript{2} New York University
 \textsuperscript{3}Princeton Language and Intelligence
\\
 \texttt{
   \{bale.chen, sw5973, cz1285\}@nyu.edu, xi.ye@princeton.edu
 }
}

\begin{document}
\maketitle
\begin{abstract}
Multi-answer question answering (QA), where questions can have many valid answers, presents a significant challenge for existing retrieval-augmented generation-based QA systems, as these systems struggle to retrieve and then synthesize a large number of evidence passages. To tackle these challenges, we propose a new multi-answer QA framework – Retrieval-augmented Independent Reading with Inter-passage Verification (RI\textsuperscript{2}VER).
Our framework retrieves a large set of passages and processes each passage individually to generate an initial high-recall but noisy answer set. Then we propose a new inter-passage verification pipeline that validates every candidate answer through (1) \textbf{Verification Question Generation}, (2) \textbf{Gathering Additional Evidence}, and (3) \textbf{Verification with inter-passage synthesis}.
Evaluations on the QAMPARI and RoMQA datasets demonstrate that our framework significantly outperforms existing baselines across various model sizes, achieving an average F1 score improvement of 11.17\%. Further analysis validates that our inter-passage verification pipeline enables our framework to be particularly beneficial for questions requiring multi-evidence synthesis.\footnote{Our code is available at \url{https://github.com/BaleChen/RIVER}}
\end{abstract}

\input{sections/intro}

\input{sections/background}

\input{sections/method}
\input{sections/experiments}
\input{sections/related-works}
\input{sections/conclusion}
\input{sections/limitations}

\bibliography{custom}

\newpage
\input{sections/appendix}

\end{document}

%% file: sections/intro.tex
\section{Introduction}

\begin{figure}[t]
    \centering
    \includegraphics[width=\columnwidth]{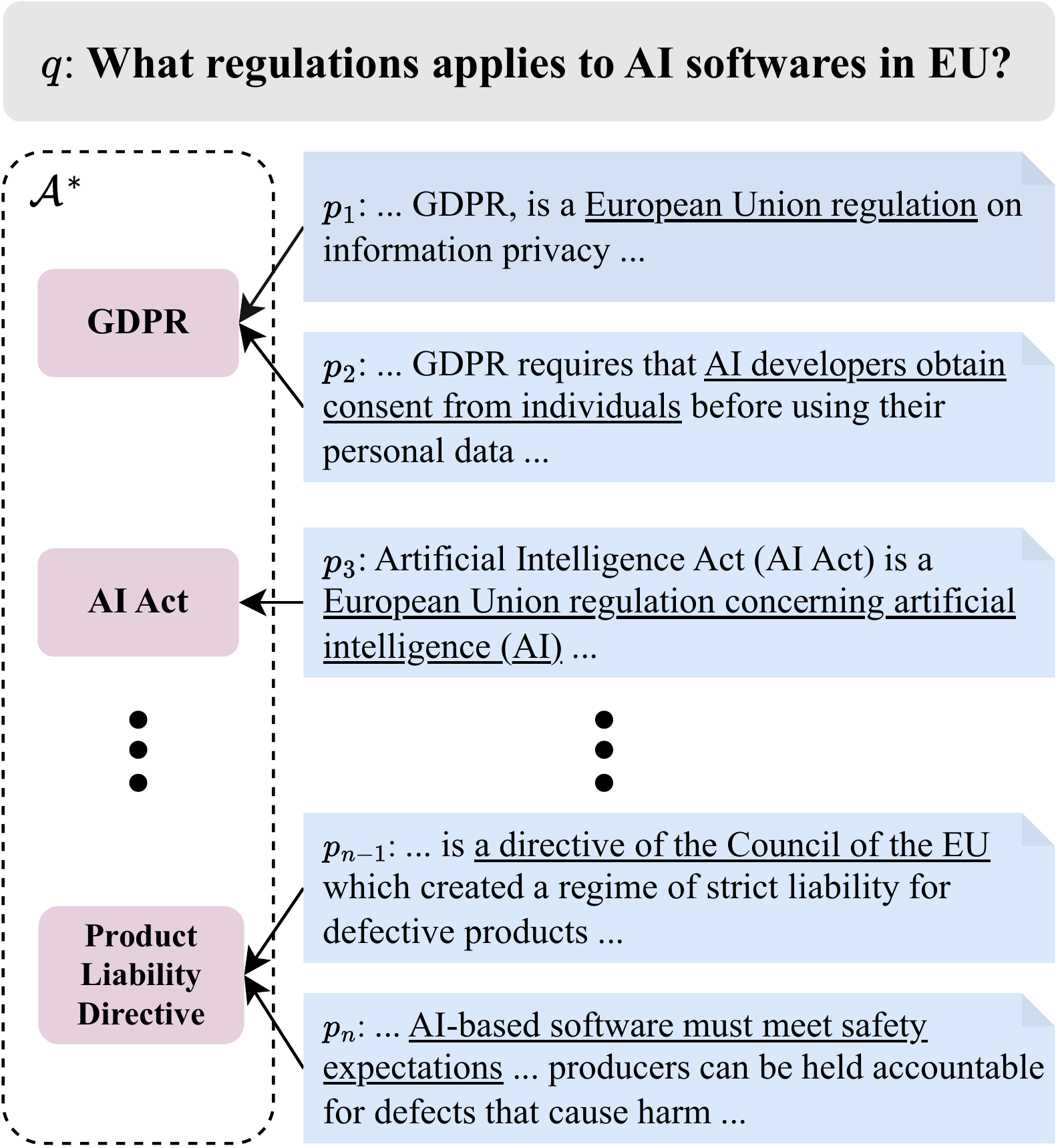}
    \caption{A representative example of a multi-evidence multi-answer question. Note that answers like "GDPR" and "Product Liability Directive" require multiple evidence passages to be supported.}
    \label{fig:example}
    \label{fig:enter-label}
\end{figure}

Question Answering (QA) represents a long-standing challenge in natural language processing \cite{Voorhees2000BuildingAQ, Chen2017ReadingWT, Min2021NeurIPS2E}, which in recent years has become an important knowledge and factuality benchmark task of large language models (LLMs) \citep{Wei2024MeasuringSF}. While most such QA benchmark datasets operate under the assumption that each question has a single answer, this assumption often fails to reflect real-world scenarios where multiple valid answers exist \citep{min2020ambigqa, amouyal-etal-2023-qampari}.
For instance, as shown in Figure~\ref{fig:example}, a legal expert assessing AI software in the EU must consider multiple applicable laws to ensure compliance, mitigate legal risks, and make informed decisions. 
Thus, developing QA systems capable of providing comprehensive information is crucial. \looseness=-1

Nevertheless, existing QA systems still fall short in providing multiple answers reliably \citep{min2020ambigqa, amouyal-etal-2023-qampari, Zhong2022RoMQAAB}. State-of-the-art LLMs, relying solely on their parametric knowledge, remain incompetent for multi-answer QA, as LLMs are prone to hallucination~\citep{Zhang2023SirensSI, Ji2022SurveyOH} and possess limited knowledge of long-tail entities~\citep{Kandpal2022LargeLM, sun-etal-2024-head}. A typical solution is to employ the retrieval-augmented generation (RAG) paradigm \citep{lewis2020rag, realm}. In such systems, a retriever module first searches for supporting passages from an external corpus based on the query, followed by the LLM that reads the passages to produce answers. However, even with external knowledge, multi-answer QA remains challenging for LLMs: First, to find all answers, LLMs need to parse information from a large number of passages. For instance, the question ``Who played midfielder for Argentina?'' has 98 correct answers, which are likely scattered across dozens of passages. Moreover, finding an answer often requires reasoning across multiple passages. For instance, in Figure~\ref{fig:example}, to correctly identify ``GDPR'' as an answer requires the LLM to synthesize two passages---one mentioning ``GDPR is an EU regulation'' and another stating ``GDPR requires AI developers to obtain user consent'' – among all other passages irrelevant to GDPR. This multi-evidence synthesis capabilities over a large amount of contextual information have been recognized as a major limitation even for modern long-context LLMs \citep{liu-etal-2024-lost, wang-etal-2024-leave, Bai2024LongBenchVT,Ye-Et-Al:2025:Longproc}. Hence, how to design QA systems with LLMs to solve those multi-evidence and multi-answer questions remains an interesting yet underexplored problem.

In this paper, we propose a new framework for multi-evidence multi-answer QA – \textbf{R}etrieval-augmented \textbf{I}ndependent Reading with \textbf{I}nter-passage \textbf{VER}ification (\textsc{RI\textsuperscript{2}VER}).
We first adopt off-the-shelf retrievers to find a large set of passages to ensure sufficient coverage. Next, an LLM processes each passage independently to extract all potentially correct candidate answers. 
This process avoids the challenge of synthesizing and reasoning over a large collection of passages but instead introduces a noisy answer candidate set, achieving high coverage of correct answers.
We then propose the inter-passage verification
(IPV) pipeline to effectively filter the noisy answer set, which is composed of three steps: 
(1) \textbf{Verification Question Generation} that 
leverages an LLM to generate verification questions about each atomic fact involved in the input question.
(2)  \textbf{Gathering Additional Evidence} that 
 collects evidence passages for each specific verification question. (3) \textbf{Verification with multi-evidence synthesis} that adopts an LLM to synthesize evidence passages and determine the correctness of the candidate answers on each verification question. Through this process, we filter out candidate answers that fail the verification steps, resulting in a refined answer set with high precision. \looseness=-1

We evaluated our framework on QAMPARI and RoMQA, two challenging multi-answer QA datasets characterized by large answer set sizes and complex questions that require multi-evidence synthesis. RI\textsuperscript{2}VER significantly outperforms baseline RAG approaches with on average an 11.17\% F1 score enhancement.
Through ablation studies, we further validate the effectiveness of \textsc{IPV}, especially for the complex subsets of questions that require multi-evidence synthesis. Our manual error analysis reveals that more than half of the prediction errors stem from suboptimal annotations. Therefore, we argue that, in addition to developing better multi-answer QA systems, ensuring more accurate answer annotations is equally crucial for the community.\looseness=-1

%% file: sections/background.tex
\section{Background: Multi-Answer QA}
We first formally define the multi-answer QA task. Given a question $q$ and a text corpus of numerous passages $\cC = \{p_i\}_{i=1}^{|C|}$, a QA system predicts a set of answers $\cA = (a_1, a_2, \dots)$ as close as the ground truth answer set $\cA^* = (a_1^*, a_2^*, \dots)$, where each answer $a_i$ is a short-form answer (e.g., an entity) to $q$. Each ground truth answer can have one or more supporting passages in $\cC$, with the latter case being multi-evidence multi-answer QA. 

The predicted answer set $\cA$ is evaluated by set precision, recall, and F1 with respect to $\cA^*$. Specifically, for each question, if a predicted answer $a_i$ exactly matches one of the ground truth answers $a_j^*$ or one of $a_j^*$'s aliases, we count it as a true positive answer, or otherwise a false positive answer. If there exists an $a_i^*$ such that neither itself nor its aliases are present in $\cA$, we count it as a false negative error. Thus, a QA system should balance between finding all correct answers (high recall) and ensuring the predicted answers are accurate (high precision) to achieve the optimal F1 score.

%% file: sections/method.tex
\begin{figure*}
    \centering
    \includegraphics[width=\textwidth]{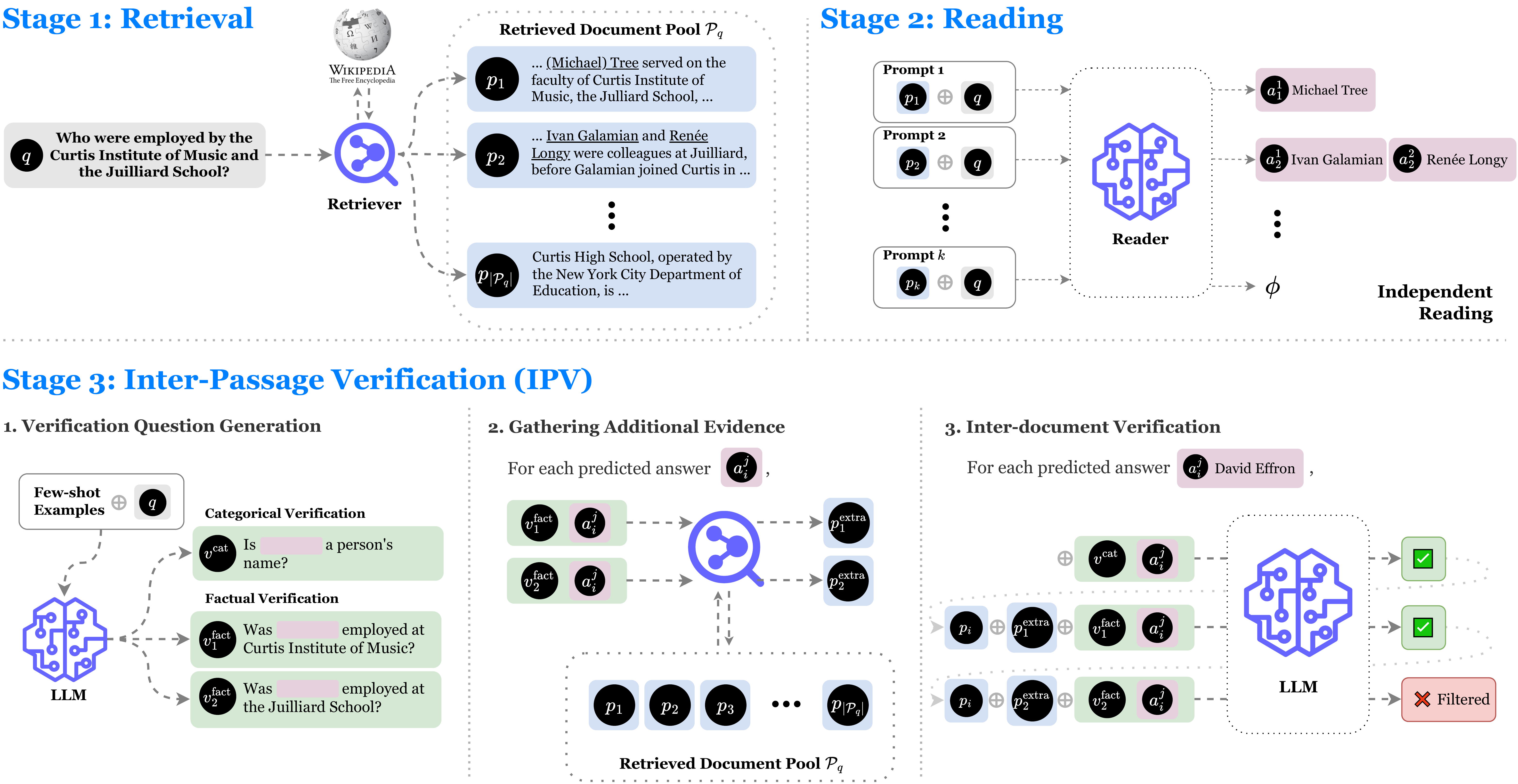}
    \caption{\textbf{Illustration of RI\textsuperscript{2}VER.} In Stage 1, we search broadly for a large pool of passages to cover sufficient evidence. In Stage 2, a reader model processes each top-$k$ retrieved passage independently to generate a set of answer candidates. In Stage 3, we employ the IPV pipeline to verify each answer against a series of verification questions generated based on the original question, along with an additional retrieval step to gather evidence.}
    \label{fig:flowchart}
\end{figure*}

\section{Method} \label{sec:method}
In this section, we introduce our multi-answer QA framework, RI\textsuperscript{2}VER, that introduces the Inter-passage Verification (IPV) stage to the vanilla RAG paradigm for multi-answer QA.
At a high level, RI\textsuperscript{2}VER first finds a large pool of passages and predicts a set of answers grounded on each passage. This retrieve-then-independently-read process prioritizes the recall of correct answers at a cost of precision, leading to many false answers in the predicted answer set. Then we filter each answer candidate to improve precision with the proposed IPV pipeline.

As illustrated in Figure~\ref{fig:flowchart}, RI\textsuperscript{2}VER is composed of three stages. First, in the retrieval stage, we broadly search for a large pool of passages. Second, in the reading stage, we employ an LLM reader with the independent reading paradigm to elicit all potential correct answers (\S \ref{sec:rag}).
Last, in the third stage, we design the \textsc{IPV} pipeline to effectively verify each answer candidate through three steps: (1) generate verification questions based on the original question (\S \ref{sec:vqg}), (2) gather additional evidence for verification by reusing the initial passage pool (\S \ref{sec:gather}), and (3) verify whether each answer candidate satisfies all verification questions, filtering the false positive answers (\S \ref{sec:veri}). \looseness=-1

\subsection{Retrieval-Augmented Independent Reading}\label{sec:rag}
RAG has demonstrated its ability to improve the language models' factuality in knowledge-intensive tasks \citep{realm, lewis2020rag}. In our framework, we adopted the retrieve-and-read paradigm with independent reading, which we illustrate in Figure~\ref{fig:flowchart} and formally describe as follows:

\paragraph{Retrieval} In the retrieval stage, we use a sparse \citep{bm25} or dense retriever \citep{karpukhin-etal-2020-dense} to retrieve a pool of passages $\cP_q = \argtopk_{p \in \cC} \texttt{Retriever}(q, p)$ that are relevant to each question $q$ from a text corpus $\cC$. $\cP_q$ is an ordered set of passages ranked by their relevance score predicted by $\texttt{Retriever}(q, p)$. 
In our \ours framework, we enlarged $|\cP_q|$ to obtain a pool of passages that ensures sufficient coverage of relevant passages about question $q$. Then, only the top-$k$ passages from the pool, denoted as $\cP_q^{\text{top-}k} = \cP_q[:k]$ are used for the reading step, while the rest of $\cP_q$ will be utilized in the subsequent \textsc{IPV} pipeline described in \S\ref{sec:vqg}-\S\ref{sec:veri}. 

\paragraph{Reading}
During the reading stage, a reader language model $\cM$ generates answers to $q$ based on the passages in $\cP_q^{\text{top-}k}$. In our framework, we adopt \textit{independent reading} where $\cM$ reads each retrieved passage $p_i \in \cP_q^{\text{top-}k}$ independently to generate answers. The reader model is instructed to generate all potentially correct candidate answers that are fully or partially supported by each passage and abstain if the passage is irrelevant to $q$. Formally, the reader model predicts a set of answers $\cM(q, p_i) = \{a_i^1, a_i^2, \dots\}$ by reading each passage $p_i$, and we take the union of generated answers from all passages, $\cA = \bigcup_{p\in\cP_q^{\text{top-}k}} \cM(q, p)$, to form the final prediction set. We denote $a_i^j$ as the $j$-th answer generated from reading passage $p_i$.

\subsection{Verification Question Generation}\label{sec:vqg}

We first generate a set of verification questions for each question $q$. The main goal of this step is to decompose complex questions into atomic factual units \citep{press-etal-2023-measuring,zhou2023leasttomost} and validate the answer for each unit. Particularly, we prompt an LLM 
to decompose the input question $q$ into a series of factual verification questions $\{v^{\text{fact}}_l\}_{l=1}^L$, with each question asking about an atomic factual unit. 
As shown in Figure~\ref{fig:flowchart}, we decompose the input question into two sub-questions, ``Was \texttt{[ANSWER]} employed at Curtis Institute of Music?'' and ``Was \texttt{[ANSWER]} employed at the Juilliard School?''. Each verification question isolates a distinct aspect of the original query, which not only makes the subsequent evidence retrieval easier but also ensures that the answer candidate satisfies all aspects of the original query.

We additionally generate a categorical verification question $v^{\text{cat}}$ for each input question that checks whether the answer is in the correct category. This design is based on the observation that LLM readers sometimes generate spurious answers that violate the category of the question (e.g. answering ``Field Concert Hall'' to ``Who were employed by the Curtis Institute of Music and the Juilliard School?'').
Both factual and categorical verification questions are generated with a placeholder (``\texttt{[ANSWER]}'') that will later be populated by answer candidates. 

\subsection{Evidence Gathering}\label{sec:gather}
Next, we gather evidence passages for each verification question. Recall that for independent reading introduced in \S\ref{sec:rag}, each predicted answer $a_i^j$ is derived from a passage $p_i$, which can be naturally used as evidence for verification. For factual verification, we conduct an extra retrieval step to obtain additional verification question-specific evidence. This extra retrieval aims to find evidence to verify the answers that require multiple evidence. For example in Figure~\ref{fig:flowchart}, $a_2^2$ = ``Renée Longy'' is generated as a potential answer from independently reading $p_2$, but verifying whether ``Renée Longy'' also ``was employed at the Curtis Institute of Music'' requires additional passages that are specific to ``Renée Longy'' and ``Curtis Institute of Music''. Thus, in this evidence-gathering step, we fill in an answer $a_i^j$ to the answer placeholder in $v^{\text{fact}}_l$, and used it as queries to retrieve $k_{\text{extra}}$ passages from $\cP_q$ using the same retriever from the retrieval stage, denoted as $\cP_l^{\text{extra}}$.\footnote{
In practice, we first do categorical verification before gathering extra evidence passages for factual verification since we can skip the extra retrieval step if the answer candidate does not pass categorical verification.} By that means, categorical verification will be grounded on $p_i$, and factual verification will be based on $\{p_i\}\cup\cP_l^{\text{extra}}$.

\subsection{Answer Verification}\label{sec:veri}
Finally, we verify each candidate answer $a_i^j$ given a sequence of verification questions $v^{\text{cat}}$ and $\{v^{\text{fact}}_l\}_{l=1}^L$ as well as the corresponding evidence passages $p_i$ or $\{p_i\}\cup\cP_l^{\text{extra}}$. 
We prompt an LLM with each verification question concatenated with its corresponding evidence passage(s) and instruct the LLM only to respond ``True'' or ``False''\footnote{We allow for minor token variations following \citeposs{Guan2023LanguageMH} implementation.}.
We compute the verification results of $a_i^j$ by comparing the probability of generating ``True'' ($p^{+}$) with ``False'' ($p^{-}$) immediately after the input prompt, and outputs ``True'' if $p^{+}$ is larger than $p^{-}$ otherwise ``False''. 
We retain an answer $a_i^j$ if the outputs for all verification questions are ``True''. \looseness=-1

%% file: sections/experiments.tex
\section{Experiments}
\subsection{Datasets}
We tested our framework on the following two multi-evidence multi-answer QA datasets. Statistics of both datasets are in Appendix~\ref{appendix:data}.

\noindent\textbf{QAMPARI}~\citep{amouyal-etal-2023-qampari} is constructed from Wikipedia's knowledge graph and tables. 
The dataset is split into simple questions (e.g., What films did Stephen King write?), intersectional questions (e.g., What films have Stephen King as screenwriter and Kubrick as director?), and compositional questions (Who is the director of a film that has Stephen King as screenwriter?). We note that the latter two types typically require more than one supporting passage for each answer. 

\noindent\textbf{RoMQA}~\citep{Zhong2022RoMQAAB} aims to test the robustness of multi-answer QA systems. It clusters entity attributes from WikiData to form complex entity-seeking queries that have many answers (Avg.~\# $= 108$) and require synthesizing multiple pieces of evidence (e.g., Which members of the Royal Society received the Order of Australia but were not employed by the University of Oxford?). 

\begin{table*}[t]
\centering
\scriptsize
\renewcommand{\arraystretch}{1.0}
\resizebox{\textwidth}{!}{%
\begin{tabular}{@{}llcccccc@{}}
\toprule
 &
   &
  \multicolumn{3}{c}{\textbf{QAMPARI}} &
  \multicolumn{3}{c}{\textbf{RoMQA}} \\ \midrule
\multicolumn{1}{l|}{} &
  \multicolumn{1}{l|}{Reader Model} &
  Prec.  &
  Rec.  &
  \multicolumn{1}{c|}{F1 ($\Delta$)} &
  Prec.  &
  Rec.  &
  F1 ($\Delta$) \\ \midrule
\multicolumn{1}{l|}{Llatrieval} &
  \multicolumn{1}{l|}{GPT-4o-mini} &
  32.04 &
  17.21 &
  \multicolumn{1}{c|}{20.29} &
  20.14 &
  15.23 &
  15.53 \\
\multicolumn{1}{l|}{Close-book} &
  \multicolumn{1}{l|}{70b} &
  27.92 &
  20.47 &
  \multicolumn{1}{c|}{21.26} &
  18.22 &
  11.53 &
  10.29 \\
\multicolumn{1}{l|}{} &
  \multicolumn{1}{l|}{GPT-4o} &
  33.98 &
  23.65 &
  \multicolumn{1}{c|}{24.59} &
  22.30 &
  12.81 &
  12.20 \\ \midrule
\multicolumn{1}{l|}{FiD} &
  \multicolumn{1}{l|}{T5} &
  36.55 &
  28.30 &
  \multicolumn{1}{c|}{30.24} &
  - &
  - &
  - \\
\multicolumn{1}{l|}{Concat.\textsuperscript{\textdagger}} &
  \multicolumn{1}{l|}{8b} &
  26.60 &
  37.84 &
  \multicolumn{1}{c|}{28.10} &
  14.27 &
  18.32 &
  10.74 \\
\multicolumn{1}{l|}{} &
  \multicolumn{1}{l|}{70b} &
  32.83 &
  37.23 &
  \multicolumn{1}{c|}{33.57} &
  21.33 &
  14.95 &
  12.24 \\
\multicolumn{1}{l|}{Indep.\textsuperscript{\textdagger}} &
  \multicolumn{1}{l|}{T5} &
  12.94 &
  32.50 &
  \multicolumn{1}{c|}{16.72} &
  - &
  - &
  - \\
\multicolumn{1}{l|}{} &
  \multicolumn{1}{l|}{8b} &
  10.24 &
  69.54 &
  \multicolumn{1}{c|}{16.28} &
  6.54 &
  44.19 &
  8.39 \\
\multicolumn{1}{l|}{} &
  \multicolumn{1}{l|}{70b} &
  27.80 &
  66.46 &
  \multicolumn{1}{c|}{35.42} &
  12.90 &
  40.17 &
  14.40 \\ \midrule
\multicolumn{1}{l|}{RI\textsuperscript{2}VER} &
  \multicolumn{1}{l|}{T5} &
  26.14  &
  28.75  &
  \multicolumn{1}{c|}{24.89 ($+$8.17)\phantom{0}} &
  - &
  - &
  - \\
\multicolumn{1}{l|}{} &
  \multicolumn{1}{l|}{8b} &
  31.41  &
  60.17  &
  \multicolumn{1}{c|}{\textbf{36.33} ($+$20.05)} &
  15.43  &
  32.08  &
  \textbf{15.83} ($+$7.44) \\
\multicolumn{1}{l|}{} &
  \multicolumn{1}{l|}{70b} &
  37.23  &
  58.91  &
  \multicolumn{1}{c|}{\textbf{40.70} ($+$5.28)\phantom{0}} &
  19.55  &
  31.53  &
  \textbf{19.24} ($+$4.84) \\ \bottomrule
\end{tabular}%
}
\caption{\textbf{Main Results.} Precision, recall, and F1 scores are macro-averaged across all test samples. The $\Delta$ values represent the F1 score difference before and after applying \textsc{IPV} on top of Independent Reading. \textsuperscript{\textdagger} Indep. = Independent Reading; Concat. = Concatenated Reading}
\label{tab:main}
\renewcommand{\arraystretch}{1}
\end{table*}

\subsection{Baselines} 
Besides the retrieve-and-read framework with \textbf{independent reading} described in Section~\ref{sec:rag}, we also compare our framework to the following baselines:

\noindent\textbf{Llatrieval} \citep{li-etal-2024-llatrieval} is a RAG system with LLM verifiers that outperforms many strong retrievers and rerankers on QAMPARI. They leveraged LLMs to verify the retrieved passage set, progressively select passages, and iteratively refine the query to improve retrieval performance. GPT-4o-mini is used for both retrieval verification and generation in the pipeline.

\noindent\textbf{Close-book QA} We also compare our methods with GPT-4 Omni \citep{Hurst2024GPT4oSC} and Llama-3.1-70b \cite{Dubey2024TheL3} in a setting without access to external passages.

\noindent\textbf{Fusion-in-Decoder} (FiD) \citep{izacard-grave-2021-leveraging} represents a retrieve-and-read baseline on QA tasks that process multiple retrieved passages in parallel through independent encoders before fusing their representations in the decoder for answer generation.

\noindent\textbf{Concatenated Reading} As an alternative to independent reading, we can instead concatenate all passages in $\cP_q^{\text{top-}k}$ alongside the question $q$ as one single prompt to the reader model. The reader model is instructed to produce a list of correct answers to $q$ based on the passages. 

\subsection{Experiment Settings}
For the retriever model, we ensemble BM25 and NV-Embed-v2\footnote{\#1 on MTEB English Retrieval Leaderboard. The rank is as of February 9, 2025, and is available \href{https://huggingface.co/spaces/mteb/leaderboard}{here}.} \citep{lee2024nv} with Reciprocal Rank Fusion \citep{rrf} for QAMPARI. On RoMQA, we only use NV-Embed-v2 for the retriever, as ensembling negatively impacts retrieval performance. We report and discuss the retriever \textsc{Answer Recall@$K$} (\textsc{ARecall@$K$}), which calculates the macro-averaged percentage of answer strings matched in the $K$ retrieved passages, in Appendix~\ref{appendix:retrieval}. We use the 2021-08-01 Wikipedia dump as the retrieval corpus $\cC$, where each Wikipedia document is chunked into 100-word passages. We initially retrieve $|\mathcal{D}_q| = 1,000$ passages for each question $q$, and the reader model reads the top $k=200$ passages. For the reader model, we use the publicly available Llama-3.1-8b-instruct and Llama-3.1-70b-instruct models, which, without ambiguity, we refer to as 8b and 70b. We also used \citeposs{amouyal-etal-2023-qampari} T5-large checkpoints for FiD and independent reading, which are finetuned on QAMPARI and Natural Questions \citep{kwiatkowski-etal-2019-natural}, thus only reported on QAMPARI. 

For our \textsc{IPV} pipeline, we first prompt 8b with few-shot examples to generate both the categorical and factual verification questions. Then, in the evidence-gathering step, we retrieve $k_{\text{extra}}=1$ extra passage from the pool $\cD_q$ for each factual verification question and discuss the impact of $k_{\mbox{extra}}$ in Section~\ref{sec:ablation}. For answer verification, we use 8b as the verifier model to compute $p^+$ and $p^-$. All prompt templates can be found in Appendix~\ref{appendix:prompt}.

\subsection{Main Results}\label{sec:results}

\noindent\textbf{Our framework substantially outperforms various baselines on both datasets.} 
As shown in the lower section of Table~\ref{tab:main}, with \ours, 8b and 70b reader models achieve an F1 score of 36.33\% and 40.70\% respectively on QAMPARI, which surpasses all other methods. 
Notably, \ours outperforms LLMs when applied in a closed-book setting (e.g., 24.59\% F1 for GPT-4o) by large margins. 
Moreover, when holding the retriever module constant, IPV also outperforms baseline retrieve-and-read methods, with the 8b model surpassing both concatenated (33.57\% F1) and independent reading (35.42\% F1) with the 70b model. We also observe consistent trends on the RoMQA dataset.

\noindent\textbf{Independent reading achieves high recall but leads to low precision.}
Independent reading enables both 8b and 70b to elicit nearly all valid answers in the retrieved passages, as evidenced by their high answer set recall (69.54\% and 66.46\% on QAMPARI) approaching the retriever's \textsc{ARecall@200} of 72.54\%.
However, especially for the 8b model, independent reading leads to low precision (10.24\% on QAMPARI and 6.54\% on RoMQA). This is likely because the 8b model is more susceptible to hallucinating on irrelevant passages when processing each passage individually, generating many false positive answers (examples in Appendix~\ref{tab:fp-example-2}). Although other methods like concatenated reading, FiD, and Llatrieval strike a better precision-recall balance, we argue that, in those frameworks, the performance is bottlenecked by the limited number of true positive answers. Instead, reading passages independently provides a favorable high-recall candidate set in \ours, since the subsequent IPV pipeline is dedicated to filtering false answers to improve precision.

\begin{figure}[t]
    \centering
    \includegraphics[width=\columnwidth]{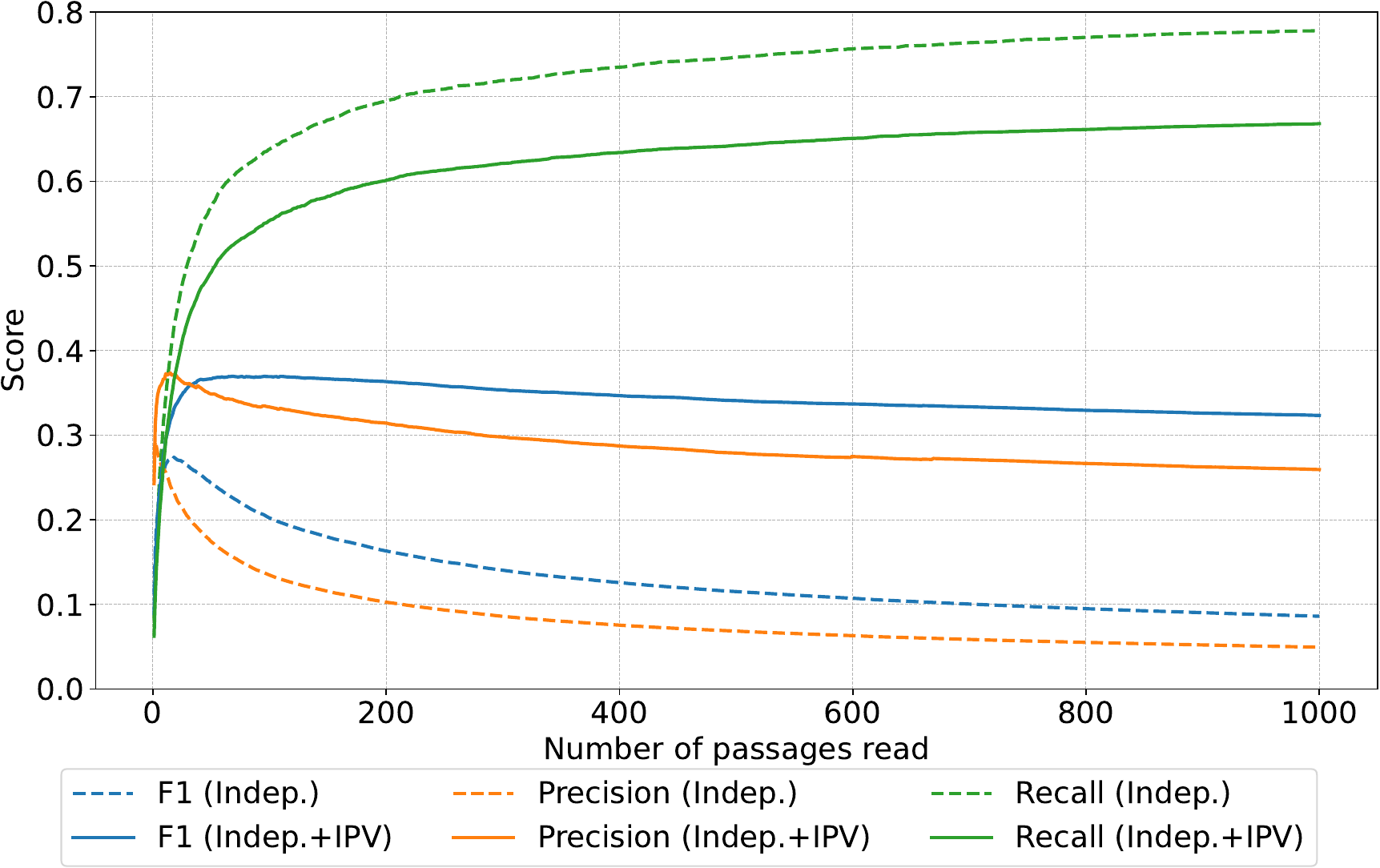}
    \caption{The prediction set precision, recall, F1 score on QAMPARI against the number of passages read by the 8b model before and after applying IPV.}
    \label{fig:4curves}
\end{figure}

\noindent\textbf{The IPV pipeline significantly improves answer set precision while maintaining high recall.} IPV effectively addresses the low precision incurred by independent reading. For QAMPARI, the precision increases by 21.17\% for 8b and 9.43\% for 70b. Similarly, on RoMQA, IPV improves precision by 8.89\% for 8b and from 6.65\% for 70b. These consistent improvements highlight IPV's ability to enhance the precision of \ours, ensuring more accurate answer sets.

Figure~\ref{fig:4curves} further demonstrates how IPV makes our framework more robust to irrelevant passages. As the dashed line indicates, 8b generates more false positive answers when reading more lower-ranked passages, leading to steeply decreasing precision. In contrast, IPV improves precision consistently over all cases and maintains relative stability as the model reads through $\cP_q$, with the 8b model's precision decreasing much more gradually and the 70b model stabilizing around 36-38\% even after reading all 1000 passages.
Notably, while recall continues to improve with reading more passages and precision stabilizes, we achieve a significantly better F1 score. This finding indicates that IPV allows \ours to benefit from the high recall of independently reading more passages without suffering from diminishing precision due to hallucinations. 
\looseness=-1

\begin{table}[t]
\centering
\resizebox{\columnwidth}{!}{%
\begin{tabular}{@{}rcccc@{}}
\toprule
\multicolumn{1}{l}{}                   & \multicolumn{3}{c}{\textbf{QAMPARI}}          & \textbf{RoMQA} \\ \cmidrule(lr){2-4}
\multicolumn{1}{l}{}                   & Simple & Inter. & Comp.                       &                \\ \midrule
\multicolumn{1}{l|}{\textsc{IPV}}               & 25.13  & 53.92  & \multicolumn{1}{l|}{41.96}  & 15.83          \\ \midrule
\multicolumn{1}{r|}{w/o Extra retrieval} & $-$1.64  & $-$3.09  & \multicolumn{1}{l|}{$-$3.10}  & $-$1.20          \\ 
\multicolumn{1}{r|}{w/o Factual Veri.}     & $-$8.73  & $-$24.12 & \multicolumn{1}{c|}{$-$18.48} & $-$5.75          \\
\multicolumn{1}{r|}{w/o Category Veri.}     & $-$1.16  & $-$0.74  & \multicolumn{1}{c|}{$-$4.30}  & $-$0.63          \\
\multicolumn{1}{r|}{Self-reflection}    & $-$1.21  & $-$8.43  & \multicolumn{1}{c|}{$-$6.45}  & $-$2.60 \\ \bottomrule         
\end{tabular}%
}
\caption{\textbf{Ablation of \textsc{IPV}.} The reader model is controlled to be 8b and we ablate each component of the IPV pipeline. The F1 scores are reported as the difference from the full IPV pipeline. Inter. = Intersectional questions, Comp. = Compositional questions.}
\label{tab:ablation}
\end{table}

\subsection{Ablation Studies}\label{sec:ablation}
In this section, we discuss the design choices and the significance of evidence gathering and verification question generation in IPV.

\begin{figure}[t]
    \centering
    \includegraphics[width=\columnwidth]{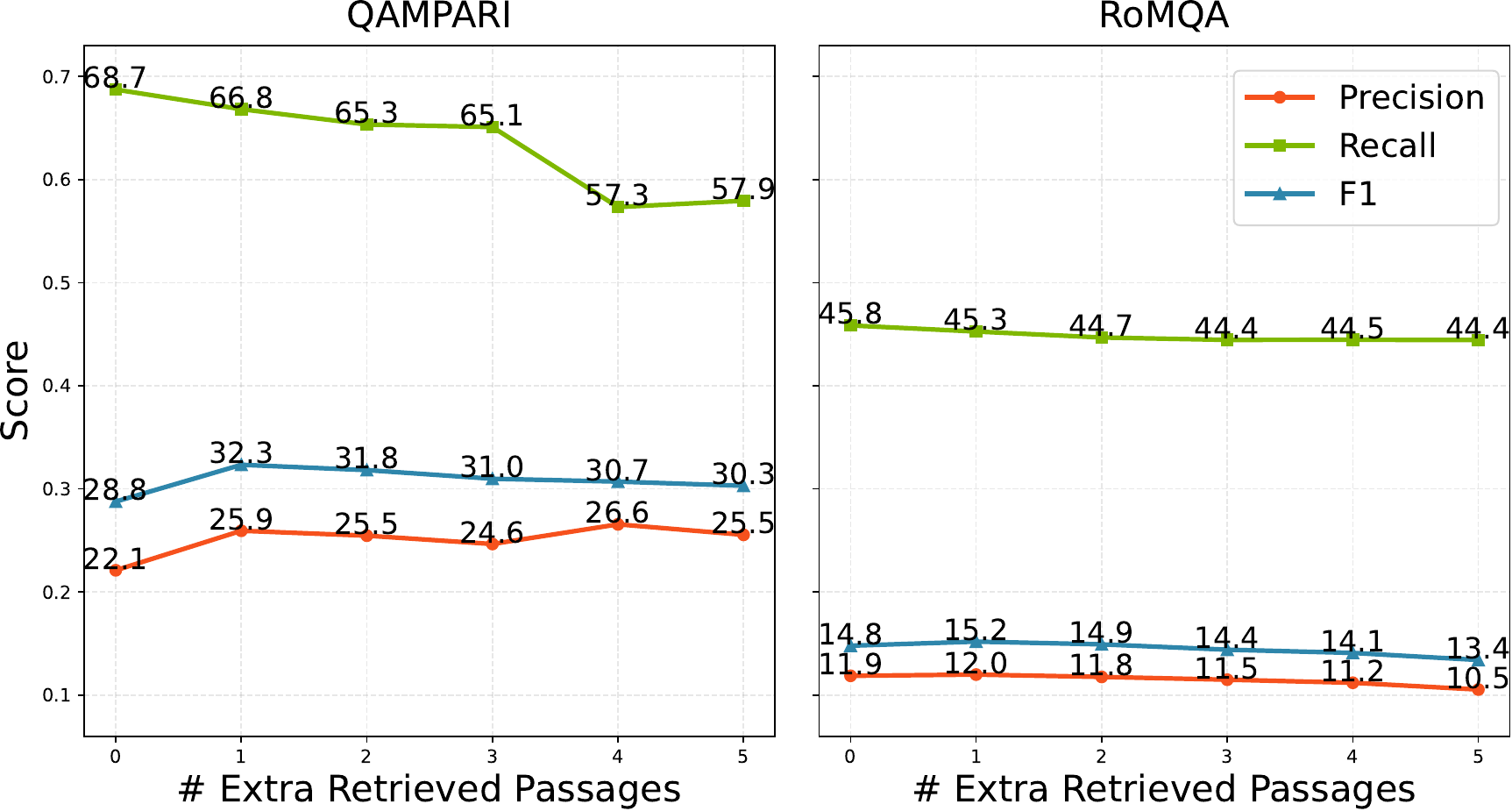}
    \caption{Precision, recall, F1 as a function of \# passages retrieved during IPV's evidence gathering step.}
    \label{fig:inter-doc-k}
\end{figure}

\noindent\textbf{Extra retrieval leads to more substantial improvements on complex questions that require evidence synthesis across passages.} As shown in Table~\ref{tab:ablation}, we compare the F1 scores between our full IPV pipeline and a simplified version that performs verification without the additional retrieval step (``w/o Extra Retrieval'') for evidence gathering. On QAMPARI, the results demonstrate that IPV's performance gains are particularly high for compositional and intersectional questions that require evidence from multiple passages to verify an answer. On RoMQA, where all questions require multiple evidence passages, IPV also shows superior results to verification with a single passage. Thus, the evidence-gathering step is especially helpful for multi-evidence, multi-answer questions.

\noindent\textbf{Optimal evidence gathering requires just one additional retrieved passage.} From Figure~\ref{fig:inter-doc-k}, we find that on both datasets, retrieving one extra passage results in better F1 scores than none or more. A potential reason is that verifying an answer might not require $\geq3$ passages, especially on QAMPARI, where the compositional and intersection questions require at most two evidence passages. Adding extra and potentially irrelevant passages distracts the model and hurts performance.

\noindent\textbf{Both factual and categorical verification are necessary in IPV.} 
We compare the contribution of factual and categorical questions in Table~\ref{tab:ablation}. We find removing either type of verification leads to performance degradation across all question types and datasets. Although most F1 improvement comes from factual verification questions, we find that checking the category of the answer first is still beneficial.  We also ablate the whole verification question generation step and reduce the answer verification to a naive self-reflection with the original question (``Self-reflection''). This leads to a significant F1 decrease, particularly profound on the intersectional and compositional questions since those questions are more challenging to verify in one run. 
\looseness=-1

\section{Error Analysis}\label{sec:error}
\begin{table}[t]
\centering
\begin{tabular}{@{}lcc@{}}
\toprule
                   & QAMPARI & RoMQA \\ \midrule
Actual FP          & 38\% & 52\% \\
Missing Annotation & 12\% & 26\% \\
Equivalent to a TP & 13\% & 6\% \\
Relatively Vague   & 5\% & 7\% \\
Relatively Specific& 32\% & 9\% \\ \bottomrule
\end{tabular}
\caption{Human-annotated error types for 100 randomly sampled false positive predictions. More than 
half of the errors stem from suboptimal 
annotations.}
\label{tab:fp-annotation}
\end{table}

\begin{table}[t]
\centering
\begin{tabular}{@{}lcc@{}}
\toprule
                      & QAMPARI & RoMQA \\ \midrule
Wrong Verification    & 12\% & 26\% \\
Insufficient evidence & 83\% & 73\% \\
Annotation Error      & 5\%  & 1\% \\ \bottomrule
\end{tabular}
\caption{Human-annotated percentage of each error type for 100 randomly sampled true positives that are filtered after IPV.}
\label{tab:fn-annotation}
\end{table}

To better understand the performance of our framework RI\textsuperscript{2}VER, we conducted an error analysis to answer the following two questions: 

\noindent\textbf{What causes the low precision on both datasets?} We randomly sample 100 false positive answers from the final predictions of 8b with IPV and manually label them into five categories shown in Table~\ref{tab:fp-annotation}. In Appendix~\ref{appendix:error}, we show an example of each type to illustrate them better. 
We find that 62\% on QAMPARI and 48\% on RoMQA of the false positive errors are not actual false positives. There is a notable amount of missing annotations or aliases of golden answers, likely because Wikipedia may contain additional answers that are not well-annotated in Wikidata, echoing the ExtendedSet results in \citet{amouyal-etal-2023-qampari}. Furthermore, we observe many false positive errors that are mistaken due to the different granularity between prediction and ground truth answers (e.g., answering ``electronics'' or ``iPhone 15'' to ``What does Apple produce?'' when the annotation is ``iPhone''). This issue in single-answer QA evaluation has been studied by \citet{Yona2024NarrowingTK}, and we advocate for future research on multi-answer QA benchmarks to annotate answers at multiple granularity levels. We also discuss the potential use of LLM-as-a-judge for evaluation in Appendix~\ref{appendix:LJ}.

\begin{table}[t]
\centering
\small
\renewcommand{\arraystretch}{1.5}
\begin{tabularx}{\columnwidth}{@{}Y{\columnwidth}@{}}
\toprule
\textbf{Question:} Paul Heller was the producer to what film? \\ \hline
\textbf{Passage:} (Title: The Magic Sword) ...Walter R. Booth as director and \underline{\smash{Robert W. Paul as producer}}. Both had worked together before with the short movie of \underline{\smash{"A Railway Collision"}} in 1900. ... \\ \hline
\textbf{Prediction:} A Railway Collision \\ \bottomrule
\end{tabularx}
\renewcommand{\arraystretch}{1.0}
\caption{A representative false positive prediction due to irrelevant context. The passage is about a different producer and film that is irrelevant to Paul Heller, while 8b hallucinates and predicts a false positive answer.}
\label{tab:fp-example-main}
\end{table}

Besides, we qualitatively inspected how the actual false positive answers are generated (Example in Table~\ref{tab:fp-example-main} and more in Table~\ref{tab:fp-example-2}). We find that they are commonly generated when the retrieved passage is irrelevant or partially relevant to the question, which concurs with the findings from \citet{Shi2023LargeLM}. In short, a significant cause of the low precision is the difficulty in multi-answer QA annotation and evaluation of varying answer granularity; yet, we acknowledge that there is still headroom for future works on developing multi-answer QA systems that are more robust to irrelevant contexts.

\noindent\textbf{Why does IPV lead to decreased recall?} We also randomly sample 100 true positive answers that are filtered during the IPV process on both datasets. We categorize each of them into three causes: wrong verification, insufficient evidence, and annotation error. 
We find that most reduction in recall comes from insufficient evidence, which means the LLM reader model produces the answer based on either partial evidence in the retrieved passage or the LLM's internal knowledge.
Since there is insufficient evidence, those answers are filtered during the IPV process, reducing the effectiveness of internal knowledge while also combating hallucinations. 
We hypothesize that IPV can further benefit from a more powerful retriever that can obtain more relevant passages both during initial retrieval and evidence gathering in IPV. 

%% file: sections/related-works.tex
\section{Related Works}
\subsection{Multi-answer QA} Many existing works on multi-answer QA focus on ambiguous questions where each interpretation can have a correct answer \citep{min2020ambigqa, stelmakh-etal-2022-asqa,lee2024ambigdocs}. For instance, several works focus on disambiguating questions and elicit more distinct and relevant answers \citep{gao-etal-2021-answering, sun-etal-2023-answering, Kim2023TreeOC}, while others tried retrieving a more diverse set of passages to cover more answers \citep{min-etal-2021-joint, In2024DiversifyverifyadaptEA, nandigam-etal-2022-diverse}. \citet{Sun2023AnsweringAQ} generates a database of disambiguated question-answer pairs to answer questions via retrieving from this database. RECTIFY \citep{Shao2021AnsweringOM} adopted a post-hoc verification pipeline, but they are limited in the scope of ambiguous questions and naive self-reflection for verification. Instead, our paper focuses on general entity-seeking questions in QAMPARI and RoMQA, which is also more challenging due to the larger answer set size and requirement for multi-passage synthesis. To our knowledge, there have not been previous works on such problems. 

\subsection{Retrieval Augmented Generation (RAG)} RAG is widely adopted in the era of LLMs since it can be directly applied to off-the-shelf LLMs or with minimal fine-tuning to reduce hallucination and augment the model's knowledge \cite{lewis2020rag, realm, Ram2023InContextRL, Shi2023REPLUGRB,gao-etal-2023-enabling,ye-etal-2024-effective}. Recent works have also explored how LLMs can in return enhance the performance of an RAG system, including refining queries \citep{ma-etal-2023-query, Chan2024RQRAGLT}, dealing with irrelevant context \citep{yoran2024making, Asai2023SelfRAGLT}, or extending to a more agentic framework that interleaves retrieval and generation \citep{Jiang2023ActiveRA, Guan2025DeepRAGTT, Singh2025AgenticRG, search-o1}. In this work, we extended the RAG framework with the IPV pipeline to solve the multi-answer QA task, significantly improving precision while maintaining high recall.

\subsection{LLM for Fact Verification} Fact verification is a well-established task in natural language understanding \citep{thorne-etal-2018-fact, Wadden2020FactOF, Honovich2022TRUERF}. With the rise of LLMs, several works have proposed to leverage LLMs' strong context understanding ability to verify factuality \citep{Zeng2023PromptTB, Guan2023LanguageMH, Tang2024MiniCheckEF}. Notably, \citet{Guan2023LanguageMH} found that relatively small LLMs are surprisingly good at verifying facts despite being prone to hallucinating in a generation. Besides, to verify complex generation beyond a single claim in an open setting, previous works have explored decomposing generation into sub-claims and then using a retriever to gather evidence from reliable sources \citep{min-etal-2023-factscore, Zhang2023TowardsLF, song-etal-2024-veriscore}. Our \textsc{IPV} pipeline borrows ideas from this line of work and effectively verifies a large set of answer candidates with relatively small-scale LLMs.

%% file: sections/conclusion.tex
\section{Conclusion}
In this paper, we propose a new multi-answer QA framework, RI\textsuperscript{2}VER, in which we retrieve broadly for relevant passages, extract all potential answers from each passage independently, and then refine the prediction set with the IPV pipeline that gathers inter-passage evidence to verify each answer candidate with an LLM. Our framework consistently outperforms various baselines across different model sizes and benchmarks. Through further analysis, we demonstrate that the \textsc{IPV} pipeline is beneficial in answering questions that require synthesizing multiple evidence passages. It also enables the reader model to benefit from the high recall from independently reading more passages without suffering from the increase in irrelevant information. We believe this work offers valuable insights and paves the way for developing more robust and reliable multi-answer QA systems. \looseness=-1

%% file: sections/limitations.tex
\section*{Limitations}

First, we did not study how the model's internal knowledge would affect the multi-answer QA performance, and instructed the model to only use evidence in context. As modern LLMs are increasingly knowledgeable, studying how to harmonize the parametric and retrieved knowledge \citep{Cheng2024UnderstandingTI} in the context of multi-answer QA is an interesting future direction. Second, RI\textsuperscript{2}VER mainly improves the answer generation and verification part, leaving the retriever module constant. As we discussed in \S\ref{sec:error}, we believe developing better retrievers in the context of multi-answer QA, potentially in the direction of diverse retrieval \citep{chen-etal-2022-design, Li2024DMQRRAGDM}, is beneficial for our and future multi-answer QA systems. Third, the IPV pipeline involves an extra retrieval step and a verification step, which introduce inference delay. We further discuss the inference latency in Appendix~\ref{appendix:latency}. Future work can explore efficient verification approaches, such as distilled retriever or verifier \citep{Zhang2024JasperAS, hsieh-etal-2023-distilling}. Lastly, the scope of our experiments is limited to Wikidata-based QA datasets and the Wikipedia corpus. Experimenting with other text corpora and developing domain-specific multi-answer QA frameworks are both promising avenues.

\section*{Acknowledgements}
Bingsen Chen, Shengjie Wang and Chen Zhao were supported by Shanghai Frontiers Science Center of Artificial Intelligence and Deep Learning, NYU Shanghai. This work was supported in part through the NYU IT High Performance Computing resources, services, and staff expertise.

%% file: sections/appendix.tex
\appendix
\section{Appendix} \label{sec:appendix}

\subsection{Retrieval Performance}
\label{appendix:retrieval}
\begin{figure}[t]
    \centering
    \includegraphics[width=\columnwidth]{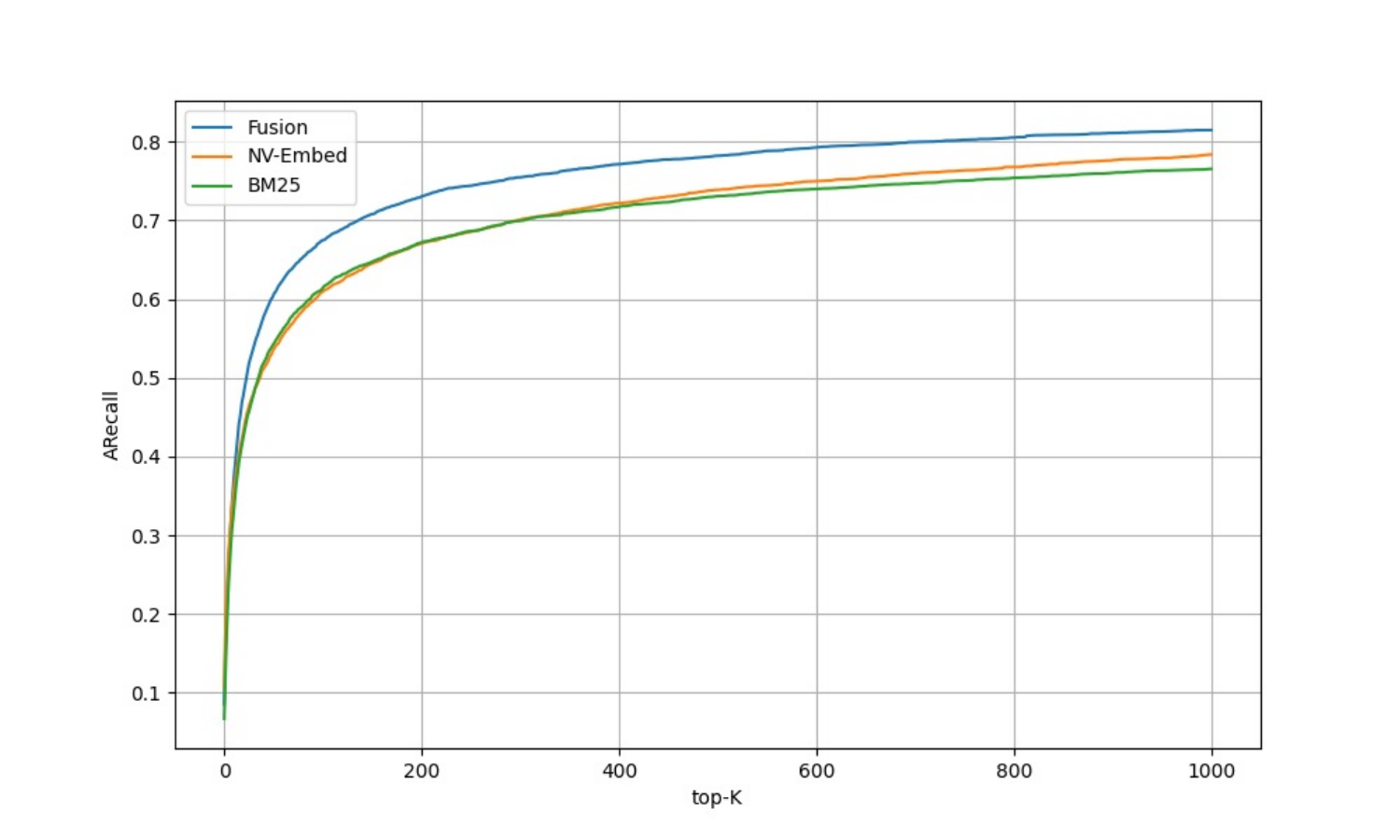}
    \caption{\textsc{ARecall@K} of BM25, NV-Embed-v2, and the fusion of BM25 and NV-Embed-v2 on QAMPARI.}
    \label{fig:retrieve-a}
\end{figure}

We evaluate the retriever performance using the \textsc{Answer Recall@K} (\textsc{ARecall@K}), which calculated the percentage of answer strings covered in the K retrieved passages macro-averaged across all test samples. In Figure~\ref{fig:retrieve-a}, we observe that on QAMPARI, ensembling the sparse and dense retriever with reciprocal rank fusion produces significantly better retrieval performance, reaching 81.23\% \textsc{ARecall@K}. Although \citet{amouyal-etal-2023-qampari} showed that BM25 is hard to beat on QAMPARI, we found that the state-of-the-art embedding model NV-Embed-v2 \citep{lee2024nv} has watched up and slightly outperformed BM25 when the number of retrieved passages is large ($|\cD_q| >400$). On RoMQA, however, we found fusing the two retrievers underperforms using NV-Embed-v2 by itself, likely because BM25 is lagging significantly behind NV-Embed-v2. \citet{Zhong2022RoMQAAB} mentioned that they found dense retrievers perform significantly more poorly than BM25 on RoMQA, but we observe that modern embedding models have outperformed BM25 by a large margin. Comparing the two datasets, we also found that RoMQA is much harder for retrievers, potentially because the questions have many more answers than QAMPARI and because the questions contain more complex requirements (e.g. ``Which members of the Royal Society received the Order of Australia, but were not employed by the University of Oxford?'').

\subsection{Examples for error analysis} \label{appendix:error}

For the false positive manual annotation, we sampled 100 false positive answers along with the corresponding question, retrieved passage, and ground truth answers for both datasets. A human annotator will verify whether this answer is indeed a false positive answer by checking whether it belongs to any of the 5 error types in Table~\ref{tab:fp-example}. The annotator has full access to the Internet. For each answer, the annotator will spend at most 5 minutes gathering information anywhere to make a decision of the 5 categories. Usually, the decision can be made very quickly with just the retrieved passage, or a simple Google search. There are a few times when the annotator cannot find enough evidence to categorize a false positive answer, even with 5 minutes, we count those as actual FPs in the statistics.

\begin{figure}[t]
    \centering
    \includegraphics[width=\columnwidth]{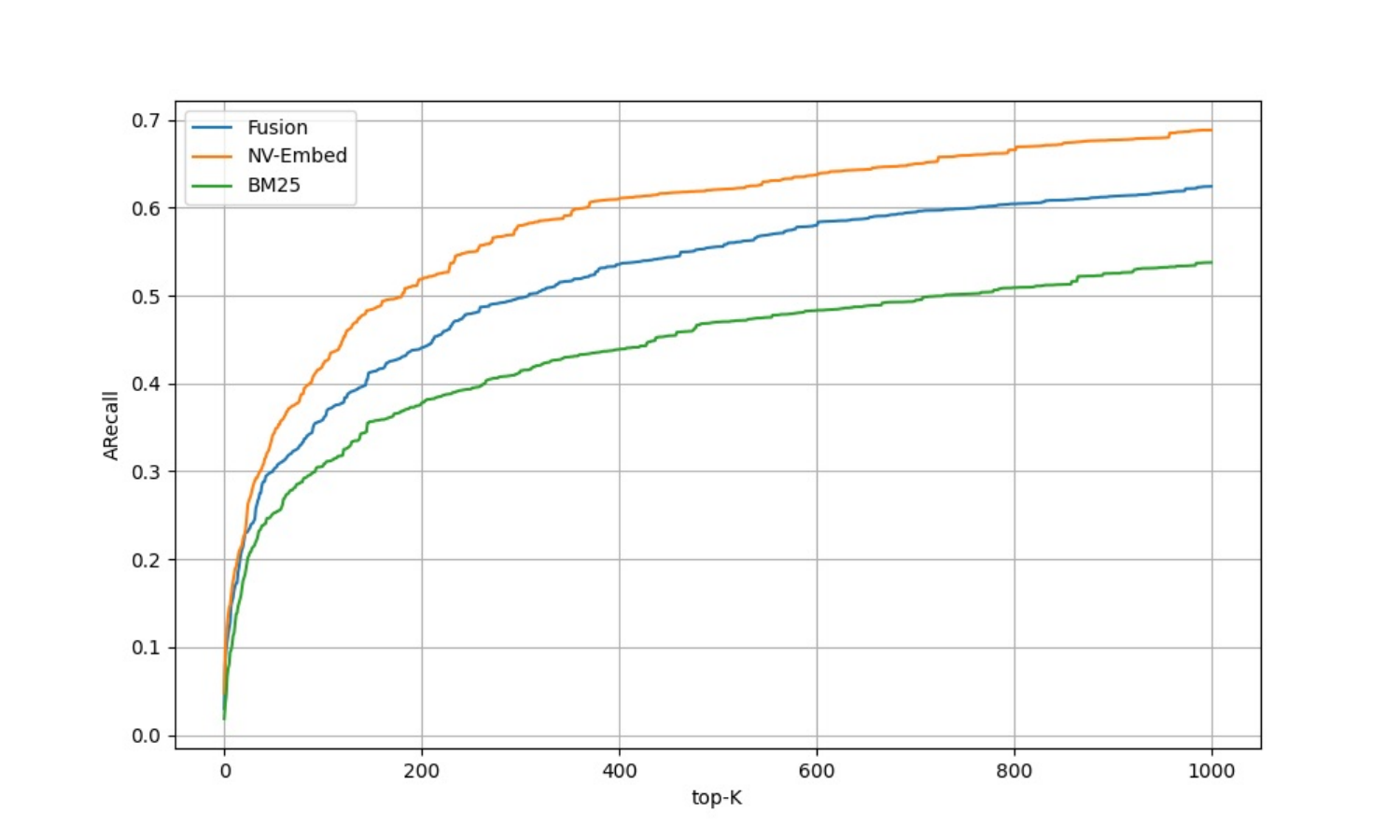}
    \caption{\textsc{ARecall@K} of BM25, NV-Embed-v2, and the fusion of BM25 and NV-Embed-v2 on QAMPARI.}
    \label{fig:retrieve-b}
\end{figure}

\begin{table}[t!]
\centering
\resizebox{\columnwidth}{!}{%
\begin{tabular}{@{}lcccc|c@{}}
\toprule
                & \multicolumn{4}{c|}{\textbf{QAMPARI}} & \multirow{2}{*}{\textbf{RoMQA}} \\ \cmidrule(lr){2-5}
 & \multicolumn{1}{l}{Simple} & \multicolumn{1}{l}{Inter.} & \multicolumn{1}{l}{Comp.} & \multicolumn{1}{l|}{All} &  \\ \midrule
\# Questions    & 500     & 200     & 300     & 1000    & 7000                            \\
Avg. \# Answers & 20      & 9       & 10      & 13      & 108                             \\ \bottomrule
\end{tabular}%
}
\caption{Data statistics of QAMPARI and RoMQA.}
\label{tab:data}
\end{table}

\subsection{Dataset statistics}\label{appendix:data}
Here we report the dataset statistics of QAMPARI and RoMQA in Table~\ref{tab:data}. For QAMPARI, we used their test split. As for RoMQA, we evaluated the development set.

\begin{table*}[t]
\centering
\small
\begin{tabularx}{\textwidth}{@{}Y{2cm}Y{3cm}Y{3cm}Y{3cm}Y{3cm}@{}}
\toprule
\textbf{Error Type}          & \textbf{Question} & \textbf{False Positive Prediction} & \textbf{Ground Truth Answers} & \textbf{Explanation} \\ \midrule
Actual FP &
  What work did Mel Brooks both write for and direct? &
  The Duchess and the Dirtwater Fox &
  Robin Hood: Men in Tights High Anxiety, History of the World, Part I, ... &
  The Duchess and the Dirtwater Fox is directed by Melvin Frank. \\ \midrule
Missing Annotation  & Who was a flute player who didn't die in Munich & Sam Most & Monty Waters, Johann Baptist Wendling, Theobald Boehm, ... & ``Heavy Metal in Baghdad'' is indeed a heavy metal film that is about a rock band. It does not exist in the ground truth set. \\ \midrule
Equivalent to a TP  & What items did Giovanola produce? &  Mésoscaphe  & Auguste Piccard, Goliath, ...    & Mésoscaphe is a missing alias of Auguste Piccard, which refers to the same product that Giovanola produced. \\ \midrule
Relatively Vague    &  What did Annibale de Gasparis discover?  &  Asteroids            &  24 Themis, 11 Parthenope, 16 Psyche, …  & Annibale de Gasparis did discover asteroids, but the annotators expect the names of specific asteroids \\ \midrule
Relatively Specific & Which examination is administered by the College Board? &  AP Statistics Exam   & Advanced Placement exams, SAT, ... &  AP statistics is one of the Advanced Placement exams, which is more granular than the ground-truth answers \\ \bottomrule
\end{tabularx}
\caption{Examples for each type of false positive error annotation.}
\label{tab:fp-example}
\end{table*}

\begin{table*}[t]
\centering
\small
\begin{tabularx}{\textwidth}{@{}Y{3cm}Y{7cm}Y{2cm}Y{3cm}@{}}
\toprule
\textbf{Question} & \textbf{Retrieved Doc.}     & \textbf{Prediction} & \textbf{Note} \\ \midrule
Who are Judge Memorial Catholic High School graduates?  &  (Title: \textbf{P. Thomas Thornbrugh}) Voters retained him again in 2016 with 61.04 percent approving. Personal. \textbf{Judge Thornbrugh} is married to Dr. Jean Thornbrugh, who is Dean of \textbf{the College for Working Adults at St. Gregory's University}. The couple have five adult children. Honors. \textbf{Thornbrugh was one of four graduates named as 2013 Distinguished Alumni by Emporia State University.} & P. Thomas Thornbrugh &  This retrieved passage is totally irrelevant to Judge Memorial Catholic High School, but there is some lexical overlap (``Judge'') and semantic overlap (graduating from a school). \\\midrule
The Arsenal F.C. won which year's FA Cup Final &  (Title: \textbf{FA Cup Final}) \textbf{This occurred on 14 occasions, the last being in 1993 between Arsenal and Sheffield Wednesday.} In September 1998, the Football Association decided that all future finals would be decided "on the day", meaning that a penalty shootout would decide the winner if the score was level after normal and extra time. Two finals since have been decided by a penalty shootout, those of 2005 (Arsenal defeating Manchester United) and 2006 (Liverpool defeating West Ham United). Stan Mortensen's hat-trick for Blackpool in 1953 is the only hat trick ever scored at Wembley in the competition's final. & 1993 & The passage never mentioned Arsenal winning the 1993 FA Cup Final. It only said that Arsenal participated in it against Sheffield Wednesday.\\ \midrule
A Tamil film starring Kamal Haasan. & (Title: Aruthu) \textbf{Aruthu is a 1976 Indian Malayalam-language film}, directed by Ravi. \textbf{The film stars Kamal Haasan}, Sumithra, M. G. Soman and Kaviyoor Ponnamma in the lead roles. The film has musical score by G. Devarajan. The film was a remake of the Hindi film \"Khel Khel Mein\". Production. \"Aruthu\" film produced by Som Prakash under production banner Sun Flower Productions. This film was shot in black-and-white. It was given an \"U\" (Unrestricted) certificate by the Central Board of Film Certification. The final length of the film was. Soundtrack. The music was composed by G. Devarajan and the lyrics were written by Yusufali Kechery. & Aruthu & The passage did mention that Aruthu features Kamal Haasan, but it is not a Tamil film.\\ \bottomrule
\end{tabularx}
\caption{Examples for actual false positive predictions with the passage from which they are generated. Relevant information is highlighted in bold font.}
\label{tab:fp-example-2}
\end{table*}

\subsection{Prompt Templates} \label{appendix:prompt}
We list all the prompt templates on the following pages. For concatenated and independent reading, the instruction part is almost the same except for some plural / single forms of words, so we show them together.

For verification question generation, we show the instructions as well as few-shot examples. Those exemplars are handpicked from the training set of both datasets. We formulate the few-shot prompting process as a multi-turn dialogue between the user and the assistant model. The role of each message is denoted \texttt{<User>} or \texttt{<Assistant>} as prefixes. On RoMQA, the instruction is slightly different because some questions might have negative constraints (e.g., What highway system located in Tottori Prefecture is \textit{\textbf{not maintained by Tottori Prefecture}}). We instruct the model to generate the verification in positive form (e.g,. Is [answer] maintained by Tottori Prefecture?) because it is empirically beneficial for the extra retrieval step and verification performance. We also make the model generate a ``tag'' (NEGATION) so that when filtering the prediction set, we filter the answer candidates that appear true to this verification question.

\subsection{Discussion on LLM-as-a-Judge for Evaluation} \label{appendix:LJ}
Due to the rigidity of exact match evaluation, many correct predictions are mistaken as incorrect, as stated in \S\ref{sec:error} and Appendix~\ref{appendix:error}. Some studies have started to use LLM-as-a-Judge \cite{zheng2023judging} to evaluate the correctness of LLM-predicted answers, allowing for format mismatch and some vocabulary variations. Since there have not been previous works on multi-answer QA that use LLM-as-a-Judge for evaluation metrics, we present a pilot implementation and analysis of our \ours framework.

\paragraph{Implementation}
We prompted GPT-4o-mini to judge the predictions on QAMPARI from concatenated reading, independent reading, and \ours, using Llama-3.1-8b-instruct as the reader model consistently. To judge each prediction, GPT-4o-mini is instructed to output the ground-truth answer index if the prediction semantically matches a ground truth, or output “None” otherwise. We then compute the set precision, recall, and F1 as usual and report in Table~\ref{tab:LG-results}.

\begin{table}[h]
\centering
\resizebox{\columnwidth}{!}{%
\begin{tabular}{l|ccc|ccc}
\toprule
& \multicolumn{3}{c|}{\textbf{Exact Match}}       & \multicolumn{3}{c}{\textbf{LLM-as-a-Judge}}     \\
& Prec.  & Rec.   & F1                  & Prec.  & Rec.   & F1                  \\
\midrule
Concat.     & 26.60  & 37.84  & 28.10               & 30.49  & 46.25  & 31.98               \\
Indep.      & 10.24  & 69.54  & 16.28               & 12.75  & 76.48  & 18.53               \\
\ours & 31.41  & 60.17  & \textbf{36.33}      & 36.74  & 68.14  & \textbf{42.73}      \\
\bottomrule
\end{tabular}%
}
\caption{Performance comparison using Exact Match and LLM-as-a-Judge for answer correctness evaluation.}
\label{tab:LG-results}
\end{table}

\paragraph{Analysis}
The LLM-as-a-judge evaluation confirms the robustness of our approach, with \ours maintaining substantial performance gains over both baseline methods. Compared to exact match evaluation, all methods demonstrate improved precision and recall under LLM-based judgment, indicating that this evaluation paradigm could better capture semantic equivalence and handle format variations. Yet, further studies are needed to verify the reliability of LLM-as-a-Judge for multi-answer QA evaluation.

While this approach effectively addresses common issues such as alias handling and format mismatches (e.g., ``Equivalent to a TP'' cases in Table~\ref{tab:fp-example}), it cannot resolve the fundamental problem of incomplete ground-truth annotations present in both datasets. Moreover, the evaluation of answers with varying granularity levels presents ongoing challenges. For instance, when the ground truth is "iPhone" but the system predicts either ``iPhone 15'' (more specific) or ``electronics'' (more general), establishing clear correctness criteria for LLM judgment remains non-trivial. 

We identify the development of refined evaluation for multi-answer QA systems as a promising direction for future research, particularly in establishing principled approaches for handling semantic granularity and annotation completeness.

\subsection{Computational Resources}
All our experiments are conducted on high-performance computing clusters. We used 1 NVIDIA A100 GPU for Llama-3.1-8b-instruct and 4 NVIDIA A100 GPUs for Llama-3.1-70b-instruct, using the vLLM \citep{kwon2023efficient} framework to speed up inference in our experiments. For retrieval, we used FAISS \citep{douze2024faiss} for efficient vector search on the large RAM (512G) CPU nodes.

\subsection{Inference Latency Analysis}\label{appendix:latency}
While our \ours framework leads to notable performance improvements, it also brings about additional inference latency. On the same computation node with 1 NVIDIA A100 GPU, we report the averaged inference wall-clock time (seconds) per question on QAMPARI with Llama-3.1-8b-instruct.

\begin{table}[h]
\centering
\small
\resizebox{\columnwidth}{!}{%
\begin{tabular}{l|c}
\toprule
Method                          & Inference Time (s) \\
\midrule
Concatenated Reading            & 9.39               \\
Independent Reading             & 6.80               \\
\ours       & 32.90              \\
\bottomrule
\end{tabular}%
}
\caption{Inference time comparison of baselines and \ours.}
\label{tab:inference_time_comparison}
\end{table}

The inference latency is mainly caused by the additional evidence gathering and the additional forward passes through the LLM to verify each answer. The former requires encoding $k$ very short verification questions and a $(1000, d) \times (d, k)$ matrix multiplication, where the number of verification questions $k$ is as small as 1 or 2, and the embedding size $d$ is 4096 in our setting. The latter step requires multiple forward passes, as many as the number of answer candidates multiplied by the number of factual verification questions. To further mitigate such latency, we discuss two potential strategies below:
\paragraph{Retrieval} Future studies can investigate using a more light-weight embedding model for the retrieval, both in the sense of a smaller embedding dimension and fewer model parameters. As research on embedding models advances, more lightweight yet performant models will improve the inference latency of our framework. Besides, embedding quantization can further squeeze the inference latency of the matrix multiplication for evidence gathering.
    
\paragraph{Verification} A more lightweight model, such as a distilled fact verification model, can potentially speed up the subsequent verification step significantly. Since the most significant bottleneck of IPV is the LLM forward passes to verify each answer candidate, a natural way to speed it up is to use a smaller model for such work. As MiniCheck~\cite{Tang2024MiniCheckEF} has shown, a similar capability can be distilled into a 770M-sized model, which means it is a viable and promising direction to accelerate the IPV pipeline with a smaller answer verifier.

\subsection{Model Details}
We here provide the specific model checkpoints used in our paper for reproduction purposes:
\begin{enumerate}
    \item Llama-3.1-8b-Instruct: \url{https://huggingface.co/meta-llama/Llama-3.1-8B-Instruct}
    \item Llama-3.1-70b-Instruct: \url{https://huggingface.co/meta-llama/Llama-3.1-70B-Instruct}
    \item NV-Embed-v2: \url{https://huggingface.co/nvidia/NV-Embed-v2}
    \item T5: \url{https://github.com/samsam3232/qampari/tree/master/models}
    \item GPT-4o: gpt-4o-2024-08-06
    \item GPT-4o-mini: gpt-4o-mini-2024-07-18
\end{enumerate}

\clearpage
\onecolumn
\begin{tcblisting}{
    colback=blue!5,    % Light blue background
    colframe=black,    % Black frame
    coltitle=white,    % White text in title
    fonttitle=\bfseries, % Bold title
    sharp corners=all, % Sharp edges
    title=Concatenated / Independent Reading,
    listing only,
    breakable,
    label=box-reading,
    listing options={
        basicstyle=\footnotesize\ttfamily,  % typewriter font
        breaklines=true,        % automatic line breaking
        breakindent=0pt,
        breakautoindent=false,
        columns=fixed,
      },
}
Read the following snippet(s) from Wikipedia documents carefully to find all the correct answers to the question. There could be multiple answers, one answer, or no answer. Do NOT generate additional descriptions/aliases/explanations! Generate the answers in the form of an unordered list as required below.

Start each answer with an asterisk and end with a new line. Your response should be in the following format if there are correct answers supported by the document snippets:

* Answer 1
* Answer 2
...

Or, if there is no correct answer, respond literally "There is no answer." without generating an unordered list.

REMINDERS:

* Please answer the question PURELY based on the documents provided. DO NOT use any additional knowledge not present in the documents.
* If the document is irrelevant to the question, do NOT use your own knowledge to answer it and just say "There is no answer".
* You should either give a list of answers as required above, or say "There is no answer". No other forms of response are accepted. 

Document Snippet(s):
{document(s)}

Question: {question} 
\end{tcblisting}
\vspace{3mm}
\begin{tcblisting}{
    colback=blue!5,    % Light blue background
    colframe=black,    % Black frame
    coltitle=white,    % White text in title
    fonttitle=\bfseries, % Bold title
    sharp corners=all, % Sharp edges
    title=Closebook,
    listing only,
    breakable,
    listing options={
        basicstyle=\footnotesize\ttfamily,  % typewriter font
        breaklines=true,        % automatic line breaking
        breakindent=0pt,
        breakautoindent=false,
        columns=fixed,
      },
}
Given a multi-answer web search question, generate ALL the answers that you know to this question. Do NOT generate additional descriptions/aliases/explanations! Please generate the answers in the form of an unordered list as required below:  

Start each answer with an asterisk and end with a new line. Your response should look like this:
* Answer 1
* Answer 2
...

Now answer this question: {question}
\end{tcblisting}

\begin{tcblisting}{
    colback=blue!5,    % Light blue background
    colframe=black,    % Black frame
    coltitle=white,    % White text in title
    fonttitle=\bfseries, % Bold title
    sharp corners=all, % Sharp edges
    title=Verification Question Generation (QAMPARI),
    listing only,
    breakable,
    listing options={
        basicstyle=\footnotesize\ttfamily,  % typewriter font
        breaklines=true,        % automatic line breaking
        breakindent=0pt,
        breakautoindent=false,
        columns=fixed,
      },
}
<User>: I am trying to answer questions that have many correct answers. I now have a set of answers for each question but some of them are incorrect and noisy. So, based on my web search questions that have many correct answers, generate 2-3 verification questions so that I can use those questions to filter my answer sets. Verification questions should always be true-or-false questions. I will retain the answer item if an answer item appears true to the question and filter out the answer items otherwise. Here are some requirements for your response:
* You should always first ask an easy category-checking question to verify whether the answer entity belongs to the correct category.
* The second or third question should purely be based on my original web search query. NO questions shall use inferred facts or external knowledge.
* Start by thinking about what questions are suitable to ask. Begin your thought with "Thought: "
* After your thought, begin asking questions after "Verification Questions: ". Start each question with a new line and an asterisk. Do NOT generate anything else after the questions.
* In the verification question, when you refer to the answer item, please leave it as [answer] and enclose it with double quotes so that I can later fill it in with items in my answer set.

<User>: My web search question: Which album has John Reuben as performer?
<Assistant>: 
Thought: To verify if an answer item is a correct album with John Reuben as the performer, the first question should confirm whether the item belongs to the category of music albums. The subsequent question should directly relate to the web search query by asking if John Reuben is credited as the performer on the album.
Verification Questions: 
* Is "[answer]" a music album?
* Is John Reuben credited as a performer on the music album "[answer]"?

<User>: My web search question: What film was directed by Radha Mohan and produced by Prakash Raj?
<Assistant>:
Thought: To filter answers effectively, the first question should confirm that the answer is a film. The second question will verify if the film was directed by Radha Mohan. The third question will confirm if the film was produced by Prakash Raj.
Verification Questions: 
* Is "[answer]" a film?
* Was the film "[answer]" directed by Radha Mohan?
* Was the film "[answer]" produced by Prakash Raj?

<User>: My web search question: Who was director of a movie penned by Christina Hodson?
<Assistant>: 
Thought: To filter answers effectively, the first question should confirm that the answer is a person. The second question will verify if there is a movie written by Christina Hodson and directed by this person.
Verification Questions: 
* Is "[answer]" a director's name?
* Is there a movie that is directed by "[answer]" and written by Christina Hodson?

User: My web search question: {question}

\end{tcblisting}

\begin{tcblisting}{
    colback=blue!5,    % Light blue background
    colframe=black,    % Black frame
    coltitle=white,    % White text in title
    fonttitle=\bfseries, % Bold title
    sharp corners=all, % Sharp edges
    title=Verification Question Generation (RoMQA),
    listing only,
    breakable,
    listing options={
        basicstyle=\footnotesize\ttfamily,  % typewriter font
        breaklines=true,        % automatic line breaking
        breakindent=0pt,
        breakautoindent=false,
        columns=fixed,
      },
}
<User> I am trying to ...

(omitted because it's the same as the instruction on QAMPARI) 

... so that I can later fill it in with items in my answer set.
* When you try to verify whether an answer does NOT have certain properties or features, ask the question in positive form and label the question with a [NEGATION] tag at the end of the question. I will retain the answers that appear false to the tagged question. 

<User> My web search question: What highway system located in Tottori Prefecture is not maintained by Tottori Prefecture
<Assistant> Thought: To verify if an answer item is a correct highway system located in Tottori Prefecture but not maintained by Tottori Prefecture, the first question should confirm whether the item belongs to the category of highway system. The subsequent questions should check whether this highway system is located in the Tottori Prefecture and meanwhile not maintained by Tottori Prefecture. I will use the [NEGATION] tag to filter out the highway systems maintained by the Tottori Prefecture.
Verification Questions: 
* Is "[answer]" a highway system?
* Is the highway system "[answer]" located in Tottori Prefecture?
* Is the highway system "[answer]" maintained by Tottori Prefecture? [NEGATION]

<User> My web search question: Which Sunni Islam figures weren't Sufist?
<Assistant> Thought: To filter answers effectively, the first question should confirm that the answer is a Sunni Islam figure. Then, the second question will verify whether this Sunni Islam figure was Sufist, which I will tag it with [NEGATION].
Verification Questions: 
* Is "[answer]" a Sunni Islam figure?
* Was the Sunni Islam figure "[answer]" Sufist? [NEGATION]

<User>My web search question: People who played for the Sheffield Wednesday F.C. and the Lincoln City F.C.
<Assistant> Thought: To filter answers effectively, the first question should confirm that the answer is a person's name. The second question will verify if this person played for Sheffield Wednesday F.C. The third question will check if this person also played for Lincoln City F.C.
Verification Questions: 
* Is "[answer]" a person's name?
* Has "[answer]" ever played for the Sheffield Wednesday F.C.?
* Has "[answer]" ever played for the Lincoln City F.C.?

<User> My web search question: {question}

\end{tcblisting}

\begin{tcblisting}{
    colback=blue!5,    % Light blue background
    colframe=black,    % Black frame
    coltitle=white,    % White text in title
    fonttitle=\bfseries, % Bold title
    sharp corners=all, % Sharp edges
    title=Verification,
    listing only,
    breakable,
    listing options={
        basicstyle=\footnotesize\ttfamily,  % typewriter font
        breaklines=true,        % automatic line breaking
        breakindent=0pt,
        breakautoindent=false,
        columns=fixed,
      },
}
Read the following document(s) carefully to answer the true-or-false question below. If the document provided is irrelevant or insufficient, then answer "False". Answer "True" if there is sufficient evidence in the document. Do not add additional description or explanation, and the answer can only be "True" or "False". Begin your response with "Answer: ...".

{documents}

Question: {question}

Answer: 

\end{tcblisting}

%% file: main.bbl
\begin{thebibliography}{69}
\providecommand{\natexlab}[1]{#1}

\bibitem[{Amouyal et~al.(2023)Amouyal, Wolfson, Rubin, Yoran, Herzig, and Berant}]{amouyal-etal-2023-qampari}
Samuel Amouyal, Tomer Wolfson, Ohad Rubin, Ori Yoran, Jonathan Herzig, and Jonathan Berant. 2023.
\newblock \href {https://aclanthology.org/2023.gem-1.9/} {{QAMPARI}: A benchmark for open-domain questions with many answers}.
\newblock In \emph{Proceedings of the Third Workshop on Natural Language Generation, Evaluation, and Metrics}.

\bibitem[{Asai et~al.(2023)Asai, Wu, Wang, Sil, and Hajishirzi}]{Asai2023SelfRAGLT}
Akari Asai, Zeqiu Wu, Yizhong Wang, Avirup Sil, and Hannaneh Hajishirzi. 2023.
\newblock \href {https://api.semanticscholar.org/CorpusID:264288947} {Self-rag: Learning to retrieve, generate, and critique through self-reflection}.
\newblock In \emph{Proceedings of the International Conference on Learning Representations}.

\bibitem[{Bai et~al.(2025)Bai, Tu, Zhang, Peng, Wang, Lv, Cao, Xu, Hou, Dong, Tang, and Li}]{Bai2024LongBenchVT}
Yushi Bai, Shangqing Tu, Jiajie Zhang, Hao Peng, Xiaozhi Wang, Xin Lv, Shulin Cao, Jiazheng Xu, Lei Hou, Yuxiao Dong, Jie Tang, and Juanzi Li. 2025.
\newblock \href {https://arxiv.org/abs/2412.15204} {Longbench v2: Towards deeper understanding and reasoning on realistic long-context multitasks}.
\newblock \emph{Preprint}, arXiv:2412.15204.

\bibitem[{Chan et~al.(2024)Chan, Xu, Yuan, Luo, Xue, Guo, and Fu}]{Chan2024RQRAGLT}
Chi-Min Chan, Chunpu Xu, Ruibin Yuan, Hongyin Luo, Wei Xue, Yike Guo, and Jie Fu. 2024.
\newblock \href {https://openreview.net/forum?id=tzE7VqsaJ4} {{RQ}-{RAG}: Learning to refine queries for retrieval augmented generation}.
\newblock In \emph{Proceedings of the Conference on Language Modeling}.

\bibitem[{Chen et~al.(2017)Chen, Fisch, Weston, and Bordes}]{Chen2017ReadingWT}
Danqi Chen, Adam Fisch, Jason Weston, and Antoine Bordes. 2017.
\newblock \href {https://doi.org/10.18653/v1/P17-1171} {Reading {W}ikipedia to answer open-domain questions}.
\newblock In \emph{Proceedings of the Association for Computational Linguistics}.

\bibitem[{Chen et~al.(2022)Chen, Liu, Uyttendaele, Zhang, Bruno, and Roth}]{chen-etal-2022-design}
Sihao Chen, Siyi Liu, Xander Uyttendaele, Yi~Zhang, William Bruno, and Dan Roth. 2022.
\newblock \href {https://doi.org/10.18653/v1/2022.findings-naacl.22} {Design challenges for a multi-perspective search engine}.
\newblock In \emph{Findings of the Association for Computational Linguistics}.

\bibitem[{Cheng et~al.(2024)Cheng, Pan, Yin, Wang, and Wang}]{Cheng2024UnderstandingTI}
Sitao Cheng, Liangming Pan, Xunjian Yin, Xinyi Wang, and William~Yang Wang. 2024.
\newblock \href {https://arxiv.org/abs/2410.08414} {Understanding the interplay between parametric and contextual knowledge for large language models}.
\newblock \emph{Preprint}, arXiv:2410.08414.

\bibitem[{Cormack et~al.(2009)Cormack, Clarke, and Buettcher}]{rrf}
Gordon~V. Cormack, Charles L~A Clarke, and Stefan Buettcher. 2009.
\newblock \href {https://doi.org/10.1145/1571941.1572114} {Reciprocal rank fusion outperforms condorcet and individual rank learning methods}.
\newblock In \emph{Proceedings of the International ACM SIGIR Conference on Research and Development in Information Retrieval}.

\bibitem[{Douze et~al.(2024)Douze, Guzhva, Deng, Johnson, Szilvasy, Mazaré, Lomeli, Hosseini, and Jégou}]{douze2024faiss}
Matthijs Douze, Alexandr Guzhva, Chengqi Deng, Jeff Johnson, Gergely Szilvasy, Pierre-Emmanuel Mazaré, Maria Lomeli, Lucas Hosseini, and Hervé Jégou. 2024.
\newblock \href {https://arxiv.org/abs/2401.08281} {The faiss library}.
\newblock \emph{Preprint}, arXiv:2401.08281.

\bibitem[{Gao et~al.(2023)Gao, Yen, Yu, and Chen}]{gao-etal-2023-enabling}
Tianyu Gao, Howard Yen, Jiatong Yu, and Danqi Chen. 2023.
\newblock \href {https://aclanthology.org/2023.emnlp-main.398.pdf} {Enabling large language models to generate text with citations}.
\newblock In \emph{Proceedings of the Conference on Empirical Methods in Natural Language Processing}.

\bibitem[{Gao et~al.(2021)Gao, Zhu, Ng, Nogueira~dos Santos, Wang, Nan, Zhang, Nallapati, Arnold, and Xiang}]{gao-etal-2021-answering}
Yifan Gao, Henghui Zhu, Patrick Ng, Cicero Nogueira~dos Santos, Zhiguo Wang, Feng Nan, Dejiao Zhang, Ramesh Nallapati, Andrew~O. Arnold, and Bing Xiang. 2021.
\newblock \href {https://doi.org/10.18653/v1/2021.acl-long.253} {Answering ambiguous questions through generative evidence fusion and round-trip prediction}.
\newblock In \emph{Proceedings of the Association for Computational Linguistics and the International Joint Conference on Natural Language Processing}.

\bibitem[{Guan et~al.(2023)Guan, Dodge, Wadden, Huang, and Peng}]{Guan2023LanguageMH}
Jian Guan, Jesse Dodge, David Wadden, Minlie Huang, and Hao Peng. 2023.
\newblock \href {https://api.semanticscholar.org/CorpusID:264426380} {Language models hallucinate, but may excel at fact verification}.
\newblock In \emph{Proceedings of the North American Chapter of the Association for Computational Linguistics}.

\bibitem[{Guan et~al.(2025)Guan, Zeng, Meng, Xin, Lu, Lin, Han, Sun, and Zhou}]{Guan2025DeepRAGTT}
Xinyan Guan, Jiali Zeng, Fandong Meng, Chunlei Xin, Yaojie Lu, Hongyu Lin, Xianpei Han, Le~Sun, and Jie Zhou. 2025.
\newblock \href {https://arxiv.org/abs/2502.01142} {Deeprag: Thinking to retrieval step by step for large language models}.
\newblock \emph{Preprint}, arXiv:2502.01142.

\bibitem[{Guu et~al.(2020)Guu, Lee, Tung, Pasupat, and Chang}]{realm}
Kelvin Guu, Kenton Lee, Zora Tung, Panupong Pasupat, and Mingwei Chang. 2020.
\newblock \href {https://proceedings.mlr.press/v119/guu20a.html} {Retrieval augmented language model pre-training}.
\newblock In \emph{Proceedings of the International Conference on Machine Learning}.

\bibitem[{Honovich et~al.(2022)Honovich, Aharoni, Herzig, Taitelbaum, Kukliansy, Cohen, Scialom, Szpektor, Hassidim, and Matias}]{Honovich2022TRUERF}
Or~Honovich, Roee Aharoni, Jonathan Herzig, Hagai Taitelbaum, Doron Kukliansy, Vered Cohen, Thomas Scialom, Idan Szpektor, Avinatan Hassidim, and Yossi Matias. 2022.
\newblock \href {https://doi.org/10.18653/v1/2022.naacl-main.287} {{TRUE}: Re-evaluating factual consistency evaluation}.
\newblock In \emph{Proceedings of the Conference of the North American Chapter of the Association for Computational Linguistics}.

\bibitem[{Hsieh et~al.(2023)Hsieh, Li, Yeh, Nakhost, Fujii, Ratner, Krishna, Lee, and Pfister}]{hsieh-etal-2023-distilling}
Cheng-Yu Hsieh, Chun-Liang Li, Chih-kuan Yeh, Hootan Nakhost, Yasuhisa Fujii, Alex Ratner, Ranjay Krishna, Chen-Yu Lee, and Tomas Pfister. 2023.
\newblock \href {https://doi.org/10.18653/v1/2023.findings-acl.507} {Distilling step-by-step! outperforming larger language models with less training data and smaller model sizes}.
\newblock In \emph{Findings of the Association for Computational Linguistics}.

\bibitem[{In et~al.(2025)In, Kim, Rossi, Tanjim, Yu, Sinha, and Park}]{In2024DiversifyverifyadaptEA}
Yeonjun In, Sungchul Kim, Ryan~A. Rossi, Mehrab Tanjim, Tong Yu, Ritwik Sinha, and Chanyoung Park. 2025.
\newblock \href {https://aclanthology.org/2025.naacl-long.56/} {Diversify-verify-adapt: Efficient and robust retrieval-augmented ambiguous question answering}.
\newblock In \emph{Proceedings of the Conference of the Nations of the Americas Chapter of the Association for Computational Linguistics}.

\bibitem[{Izacard and Grave(2021)}]{izacard-grave-2021-leveraging}
Gautier Izacard and Edouard Grave. 2021.
\newblock \href {https://doi.org/10.18653/v1/2021.eacl-main.74} {Leveraging passage retrieval with generative models for open domain question answering}.
\newblock In \emph{Proceedings of the Conference of the European Chapter of the Association for Computational Linguistics}.

\bibitem[{Ji et~al.(2022)Ji, Lee, Frieske, Yu, Su, Xu, Ishii, Bang, Chen, Dai, Madotto, and Fung}]{Ji2022SurveyOH}
Ziwei Ji, Nayeon Lee, Rita Frieske, Tiezheng Yu, Dan Su, Yan Xu, Etsuko Ishii, Yejin Bang, Delong Chen, Wenliang Dai, Andrea Madotto, and Pascale Fung. 2022.
\newblock \href {https://api.semanticscholar.org/CorpusID:246652372} {Survey of hallucination in natural language generation}.
\newblock \emph{ACM Computing Surveys}.

\bibitem[{Jiang et~al.(2023)Jiang, Xu, Gao, Sun, Liu, Dwivedi-Yu, Yang, Callan, and Neubig}]{Jiang2023ActiveRA}
Zhengbao Jiang, Frank~F. Xu, Luyu Gao, Zhiqing Sun, Qian Liu, Jane Dwivedi-Yu, Yiming Yang, Jamie Callan, and Graham Neubig. 2023.
\newblock \href {https://arxiv.org/abs/2305.06983} {Active retrieval augmented generation}.
\newblock \emph{Preprint}, arXiv:2305.06983.

\bibitem[{Kandpal et~al.(2022)Kandpal, Deng, Roberts, Wallace, and Raffel}]{Kandpal2022LargeLM}
Nikhil Kandpal, H.~Deng, Adam Roberts, Eric Wallace, and Colin Raffel. 2022.
\newblock \href {https://api.semanticscholar.org/CorpusID:253522998} {Large language models struggle to learn long-tail knowledge}.
\newblock In \emph{Proceedings of the International Conference on Machine Learning}.

\bibitem[{Karpukhin et~al.(2020)Karpukhin, Oguz, Min, Lewis, Wu, Edunov, Chen, and Yih}]{karpukhin-etal-2020-dense}
Vladimir Karpukhin, Barlas Oguz, Sewon Min, Patrick Lewis, Ledell Wu, Sergey Edunov, Danqi Chen, and Wen-tau Yih. 2020.
\newblock \href {https://doi.org/10.18653/v1/2020.emnlp-main.550} {Dense passage retrieval for open-domain question answering}.
\newblock In \emph{Proceedings of the Conference on Empirical Methods in Natural Language Processing}.

\bibitem[{Kim et~al.(2023)Kim, Kim, Jeon, Park, and Kang}]{Kim2023TreeOC}
Gangwoo Kim, Sungdong Kim, Byeongguk Jeon, Joonsuk Park, and Jaewoo Kang. 2023.
\newblock \href {https://api.semanticscholar.org/CorpusID:264426402} {Tree of clarifications: Answering ambiguous questions with retrieval-augmented large language models}.
\newblock In \emph{Proceedings of the Conference on Empirical Methods in Natural Language Processing}.

\bibitem[{Kwiatkowski et~al.(2019)Kwiatkowski, Palomaki, Redfield, Collins, Parikh, Alberti, Epstein, Polosukhin, Devlin, Lee, Toutanova, Jones, Kelcey, Chang, Dai, Uszkoreit, Le, and Petrov}]{kwiatkowski-etal-2019-natural}
Tom Kwiatkowski, Jennimaria Palomaki, Olivia Redfield, Michael Collins, Ankur Parikh, Chris Alberti, Danielle Epstein, Illia Polosukhin, Jacob Devlin, Kenton Lee, Kristina Toutanova, Llion Jones, Matthew Kelcey, Ming-Wei Chang, Andrew~M. Dai, Jakob Uszkoreit, Quoc Le, and Slav Petrov. 2019.
\newblock \href {https://doi.org/10.1162/tacl_a_00276} {Natural questions: A benchmark for question answering research}.
\newblock \emph{Transactions of the Association for Computational Linguistics}.

\bibitem[{Kwon et~al.(2023)Kwon, Li, Zhuang, Sheng, Zheng, Yu, Gonzalez, Zhang, and Stoica}]{kwon2023efficient}
Woosuk Kwon, Zhuohan Li, Siyuan Zhuang, Ying Sheng, Lianmin Zheng, Cody~Hao Yu, Joseph~E. Gonzalez, Hao Zhang, and Ion Stoica. 2023.
\newblock \href {https://arxiv.org/abs/2309.06180} {Efficient memory management for large language model serving with pagedattention}.
\newblock In \emph{Proceedings of the ACM Symposium on Operating Systems Principles}.

\bibitem[{Lee et~al.(2025)Lee, Roy, Xu, Raiman, Shoeybi, Catanzaro, and Ping}]{lee2024nv}
Chankyu Lee, Rajarshi Roy, Mengyao Xu, Jonathan Raiman, Mohammad Shoeybi, Bryan Catanzaro, and Wei Ping. 2025.
\newblock \href {https://arxiv.org/abs/2405.17428} {Nv-embed: Improved techniques for training llms as generalist embedding models}.
\newblock \emph{Preprint}, arXiv:2405.17428.

\bibitem[{Lee et~al.(2024)Lee, Ye, and Choi}]{lee2024ambigdocs}
Yoonsang Lee, Xi~Ye, and Eunsol Choi. 2024.
\newblock \href {https://openreview.net/forum?id=mkYCfO822n} {Ambigdocs: Reasoning across documents on different entities under the same name}.
\newblock In \emph{Proceedings of the Conference on Language Modeling}.

\bibitem[{Lewis et~al.(2020)Lewis, Perez, Piktus, Petroni, Karpukhin, Goyal, K\"{u}ttler, Lewis, Yih, Rockt\"{a}schel, Riedel, and Kiela}]{lewis2020rag}
Patrick Lewis, Ethan Perez, Aleksandra Piktus, Fabio Petroni, Vladimir Karpukhin, Naman Goyal, Heinrich K\"{u}ttler, Mike Lewis, Wen-tau Yih, Tim Rockt\"{a}schel, Sebastian Riedel, and Douwe Kiela. 2020.
\newblock \href {https://proceedings.neurips.cc/paper_files/paper/2020/file/6b493230205f780e1bc26945df7481e5-Paper.pdf} {Retrieval-augmented generation for knowledge-intensive nlp tasks}.
\newblock In \emph{Proceedings of the Advances in Neural Information Processing Systems}.

\bibitem[{Li et~al.(2024{\natexlab{a}})Li, Zhu, Li, Yin, Sun, and Qiu}]{li-etal-2024-llatrieval}
Xiaonan Li, Changtai Zhu, Linyang Li, Zhangyue Yin, Tianxiang Sun, and Xipeng Qiu. 2024{\natexlab{a}}.
\newblock \href {https://aclanthology.org/2024.naacl-long.305} {{LL}atrieval: {LLM}-verified retrieval for verifiable generation}.
\newblock In \emph{Proceedings of the Conference of the North American Chapter of the Association for Computational Linguistics}.

\bibitem[{Li et~al.(2025)Li, Dong, Jin, Zhang, Zhou, Zhu, Zhang, and Dou}]{search-o1}
Xiaoxi Li, Guanting Dong, Jiajie Jin, Yuyao Zhang, Yujia Zhou, Yutao Zhu, Peitian Zhang, and Zhicheng Dou. 2025.
\newblock \href {https://arxiv.org/abs/2501.05366} {Search-o1: Agentic search-enhanced large reasoning models}.
\newblock \emph{Preprint}, arXiv:2501.05366.

\bibitem[{Li et~al.(2024{\natexlab{b}})Li, Wang, Jiang, Mao, Chen, Du, Zhang, Zhang, Zhang, and Liu}]{Li2024DMQRRAGDM}
Zhicong Li, Jiahao Wang, Zhishu Jiang, Hangyu Mao, Zhongxia Chen, Jiazhen Du, Yuanxing Zhang, Fuzheng Zhang, Di~Zhang, and Yong Liu. 2024{\natexlab{b}}.
\newblock \href {https://arxiv.org/abs/2411.13154} {Dmqr-rag: Diverse multi-query rewriting for rag}.
\newblock \emph{Preprint}, arXiv:2411.13154.

\bibitem[{Liu et~al.(2024)Liu, Lin, Hewitt, Paranjape, Bevilacqua, Petroni, and Liang}]{liu-etal-2024-lost}
Nelson~F. Liu, Kevin Lin, John Hewitt, Ashwin Paranjape, Michele Bevilacqua, Fabio Petroni, and Percy Liang. 2024.
\newblock \href {https://doi.org/10.1162/tacl_a_00638} {Lost in the middle: How language models use long contexts}.
\newblock \emph{Transactions of the Association for Computational Linguistics}.

\bibitem[{Ma et~al.(2023)Ma, Gong, He, Zhao, and Duan}]{ma-etal-2023-query}
Xinbei Ma, Yeyun Gong, Pengcheng He, Hai Zhao, and Nan Duan. 2023.
\newblock \href {https://doi.org/10.18653/v1/2023.emnlp-main.322} {Query rewriting in retrieval-augmented large language models}.
\newblock In \emph{Proceedings of the Conference on Empirical Methods in Natural Language Processing}.

\bibitem[{Min et~al.(2021{\natexlab{a}})Min, Boyd-Graber, Alberti, Chen, Choi, Collins, Guu, Hajishirzi, Lee, Palomaki, Raffel, Roberts, Kwiatkowski, Lewis, Wu, K\"uttler, Liu, Minervini, Stenetorp, Riedel, Yang, Seo, Izacard, Petroni, Hosseini, Cao, Grave, Yamada, Shimaoka, Suzuki, Miyawaki, Sato, Takahashi, Suzuki, Fajcik, Docekal, Ondrej, Smrz, Cheng, Shen, Liu, He, Chen, Gao, Oguz, Chen, Karpukhin, Peshterliev, Okhonko, Schlichtkrull, Gupta, Mehdad, and Yih}]{Min2021NeurIPS2E}
Sewon Min, Jordan Boyd-Graber, Chris Alberti, Danqi Chen, Eunsol Choi, Michael Collins, Kelvin Guu, Hannaneh Hajishirzi, Kenton Lee, Jennimaria Palomaki, Colin Raffel, Adam Roberts, Tom Kwiatkowski, Patrick Lewis, Yuxiang Wu, Heinrich K\"uttler, Linqing Liu, Pasquale Minervini, Pontus Stenetorp, Sebastian Riedel, Sohee Yang, Minjoon Seo, Gautier Izacard, Fabio Petroni, Lucas Hosseini, Nicola~De Cao, Edouard Grave, Ikuya Yamada, Sonse Shimaoka, Masatoshi Suzuki, Shumpei Miyawaki, Shun Sato, Ryo Takahashi, Jun Suzuki, Martin Fajcik, Martin Docekal, Karel Ondrej, Pavel Smrz, Hao Cheng, Yelong Shen, Xiaodong Liu, Pengcheng He, Weizhu Chen, Jianfeng Gao, Barlas Oguz, Xilun Chen, Vladimir Karpukhin, Stan Peshterliev, Dmytro Okhonko, Michael Schlichtkrull, Sonal Gupta, Yashar Mehdad, and Wen-tau Yih. 2021{\natexlab{a}}.
\newblock \href {https://proceedings.mlr.press/v133/min21a.html} {Neurips 2020 efficientqa competition: Systems, analyses and lessons learned}.
\newblock In \emph{Proceedings of the NeurIPS 2020 Competition and Demonstration Track}.

\bibitem[{Min et~al.(2023)Min, Krishna, Lyu, Lewis, Yih, Koh, Iyyer, Zettlemoyer, and Hajishirzi}]{min-etal-2023-factscore}
Sewon Min, Kalpesh Krishna, Xinxi Lyu, Mike Lewis, Wen-tau Yih, Pang Koh, Mohit Iyyer, Luke Zettlemoyer, and Hannaneh Hajishirzi. 2023.
\newblock \href {https://doi.org/10.18653/v1/2023.emnlp-main.741} {{FA}ct{S}core: Fine-grained atomic evaluation of factual precision in long form text generation}.
\newblock In \emph{Proceedings of the Conference on Empirical Methods in Natural Language Processing}.

\bibitem[{Min et~al.(2021{\natexlab{b}})Min, Lee, Chang, Toutanova, and Hajishirzi}]{min-etal-2021-joint}
Sewon Min, Kenton Lee, Ming-Wei Chang, Kristina Toutanova, and Hannaneh Hajishirzi. 2021{\natexlab{b}}.
\newblock \href {https://doi.org/10.18653/v1/2021.emnlp-main.560} {Joint passage ranking for diverse multi-answer retrieval}.
\newblock In \emph{Proceedings of the Conference on Empirical Methods in Natural Language Processing}.

\bibitem[{Min et~al.(2020)Min, Michael, Hajishirzi, and Zettlemoyer}]{min2020ambigqa}
Sewon Min, Julian Michael, Hannaneh Hajishirzi, and Luke Zettlemoyer. 2020.
\newblock \href {https://doi.org/10.18653/v1/2020.emnlp-main.466} {{A}mbig{QA}: Answering ambiguous open-domain questions}.
\newblock In \emph{Proceedings of the Conference on Empirical Methods in Natural Language Processing}.

\bibitem[{Nandigam et~al.(2022)Nandigam, Rayaprolu, and Shrivastava}]{nandigam-etal-2022-diverse}
Poojitha Nandigam, Nikhil Rayaprolu, and Manish Shrivastava. 2022.
\newblock \href {https://aclanthology.org/2022.coling-1.194/} {Diverse multi-answer retrieval with determinantal point processes}.
\newblock In \emph{Proceedings of the 29th International Conference on Computational Linguistics}.

\bibitem[{OpenAI(2024)}]{Hurst2024GPT4oSC}
OpenAI. 2024.
\newblock \href {https://arxiv.org/abs/2410.21276} {Gpt-4o system card}.
\newblock \emph{Preprint}, arXiv:2410.21276.

\bibitem[{Press et~al.(2023)Press, Zhang, Min, Schmidt, Smith, and Lewis}]{press-etal-2023-measuring}
Ofir Press, Muru Zhang, Sewon Min, Ludwig Schmidt, Noah Smith, and Mike Lewis. 2023.
\newblock \href {https://doi.org/10.18653/v1/2023.findings-emnlp.378} {Measuring and narrowing the compositionality gap in language models}.
\newblock In \emph{Findings of the Association for Computational Linguistics}.

\bibitem[{Ram et~al.(2023)Ram, Levine, Dalmedigos, Muhlgay, Shashua, Leyton-Brown, and Shoham}]{Ram2023InContextRL}
Ori Ram, Yoav Levine, Itay Dalmedigos, Dor Muhlgay, Amnon Shashua, Kevin Leyton-Brown, and Yoav Shoham. 2023.
\newblock \href {https://api.semanticscholar.org/CorpusID:256459451} {In-context retrieval-augmented language models}.
\newblock \emph{Transactions of the Association for Computational Linguistics}.

\bibitem[{Robertson and Zaragoza(2009)}]{bm25}
Stephen Robertson and Hugo Zaragoza. 2009.
\newblock \href {https://doi.org/10.1561/1500000019} {The probabilistic relevance framework: Bm25 and beyond}.
\newblock \emph{Foundations and Trends in Information Retrieval}.

\bibitem[{Shao and Huang(2021)}]{Shao2021AnsweringOM}
Zhihong Shao and Minlie Huang. 2021.
\newblock \href {https://api.semanticscholar.org/CorpusID:247188139} {Answering open-domain multi-answer questions via a recall-then-verify framework}.
\newblock In \emph{Proceedings of the Association for Computational Linguistics}.

\bibitem[{Shi et~al.(2023{\natexlab{a}})Shi, Chen, Misra, Scales, Dohan, Chi, Scharli, and Zhou}]{Shi2023LargeLM}
Freda Shi, Xinyun Chen, Kanishka Misra, Nathan Scales, David Dohan, Ed~H. Chi, Nathanael Scharli, and Denny Zhou. 2023{\natexlab{a}}.
\newblock \href {https://api.semanticscholar.org/CorpusID:256459776} {Large language models can be easily distracted by irrelevant context}.
\newblock In \emph{Proceedings of the International Conference on Machine Learning}.

\bibitem[{Shi et~al.(2023{\natexlab{b}})Shi, Min, Yasunaga, Seo, James, Lewis, Zettlemoyer, and tau Yih}]{Shi2023REPLUGRB}
Weijia Shi, Sewon Min, Michihiro Yasunaga, Minjoon Seo, Rich James, Mike Lewis, Luke Zettlemoyer, and Wen tau Yih. 2023{\natexlab{b}}.
\newblock \href {https://api.semanticscholar.org/CorpusID:256389797} {Replug: Retrieval-augmented black-box language models}.
\newblock In \emph{Proceedings of the North American Chapter of the Association for Computational Linguistics}.

\bibitem[{Singh et~al.(2025)Singh, Ehtesham, Kumar, and Khoei}]{Singh2025AgenticRG}
Aditi Singh, Abul Ehtesham, Saket Kumar, and Tala~Talaei Khoei. 2025.
\newblock \href {https://arxiv.org/abs/2501.09136} {Agentic retrieval-augmented generation: A survey on agentic rag}.
\newblock \emph{Preprint}, arXiv:2501.09136.

\bibitem[{Song et~al.(2024)Song, Kim, and Iyyer}]{song-etal-2024-veriscore}
Yixiao Song, Yekyung Kim, and Mohit Iyyer. 2024.
\newblock \href {https://doi.org/10.18653/v1/2024.findings-emnlp.552} {{V}eri{S}core: Evaluating the factuality of verifiable claims in long-form text generation}.
\newblock In \emph{Findings of the Association for Computational Linguistics}.

\bibitem[{Stelmakh et~al.(2022)Stelmakh, Luan, Dhingra, and Chang}]{stelmakh-etal-2022-asqa}
Ivan Stelmakh, Yi~Luan, Bhuwan Dhingra, and Ming-Wei Chang. 2022.
\newblock \href {https://doi.org/10.18653/v1/2022.emnlp-main.566} {{ASQA}: Factoid questions meet long-form answers}.
\newblock In \emph{Proceedings of the Conference on Empirical Methods in Natural Language Processing}.

\bibitem[{Sun et~al.(2023{\natexlab{a}})Sun, Cohen, and Salakhutdinov}]{Sun2023AnsweringAQ}
Haitian Sun, William~W. Cohen, and Ruslan Salakhutdinov. 2023{\natexlab{a}}.
\newblock \href {https://arxiv.org/abs/2308.08661} {Answering ambiguous questions with a database of questions, answers, and revisions}.
\newblock \emph{Preprint}, arXiv:2308.08661.

\bibitem[{Sun et~al.(2024)Sun, Xu, Zha, Liu, and Dong}]{sun-etal-2024-head}
Kai Sun, Yifan Xu, Hanwen Zha, Yue Liu, and Xin~Luna Dong. 2024.
\newblock \href {https://doi.org/10.18653/v1/2024.naacl-long.18} {Head-to-tail: How knowledgeable are large language models ({LLM}s)? {A}.{K}.{A}. will {LLM}s replace knowledge graphs?}
\newblock In \emph{Proceedings of the Conference of the North American Chapter of the Association for Computational Linguistics}.

\bibitem[{Sun et~al.(2023{\natexlab{b}})Sun, Cai, Chen, Ren, Chen, de~Rijke, and Ren}]{sun-etal-2023-answering}
Weiwei Sun, Hengyi Cai, Hongshen Chen, Pengjie Ren, Zhumin Chen, Maarten de~Rijke, and Zhaochun Ren. 2023{\natexlab{b}}.
\newblock \href {https://doi.org/10.18653/v1/2023.acl-long.424} {Answering ambiguous questions via iterative prompting}.
\newblock In \emph{Proceedings of the Association for Computational Linguistics}.

\bibitem[{Tang et~al.(2024)Tang, Laban, and Durrett}]{Tang2024MiniCheckEF}
Liyan Tang, Philippe Laban, and Greg Durrett. 2024.
\newblock \href {https://api.semanticscholar.org/CorpusID:269157443} {Minicheck: Efficient fact-checking of llms on grounding documents}.
\newblock In \emph{Proceedings of the Conference on Empirical Methods in Natural Language Processing}.

\bibitem[{Team(2024)}]{Dubey2024TheL3}
Meta~Llama Team. 2024.
\newblock \href {https://arxiv.org/abs/2407.21783} {The llama 3 herd of models}.
\newblock \emph{Preprint}, arXiv:2407.21783.

\bibitem[{Thorne et~al.(2018)Thorne, Vlachos, Cocarascu, Christodoulopoulos, and Mittal}]{thorne-etal-2018-fact}
James Thorne, Andreas Vlachos, Oana Cocarascu, Christos Christodoulopoulos, and Arpit Mittal. 2018.
\newblock \href {https://doi.org/10.18653/v1/W18-5501} {The fact extraction and {VER}ification ({FEVER}) shared task}.
\newblock In \emph{Proceedings of the First Workshop on Fact Extraction and {VER}ification}.

\bibitem[{Voorhees and Tice(2000)}]{Voorhees2000BuildingAQ}
Ellen~M. Voorhees and Dawn~M. Tice. 2000.
\newblock \href {https://api.semanticscholar.org/CorpusID:11465263} {Building a question answering test collection}.
\newblock In \emph{Proceedings of the International ACM SIGIR Conference on Research and Development in Information Retrieval}.

\bibitem[{Wadden et~al.(2020)Wadden, Lo, Wang, Lin, van Zuylen, Cohan, and Hajishirzi}]{Wadden2020FactOF}
David Wadden, Kyle Lo, Lucy~Lu Wang, Shanchuan Lin, Madeleine van Zuylen, Arman Cohan, and Hannaneh Hajishirzi. 2020.
\newblock \href {https://api.semanticscholar.org/CorpusID:216867133} {Fact or fiction: Verifying scientific claims}.
\newblock In \emph{Proceedings of the Conference on Empirical Methods in Natural Language Processing}.

\bibitem[{Wang et~al.(2024)Wang, Chen, Cheng, Liao, Zhang, Wu, Yu, Xu, Zhang, Luo, Li, Yang, Huang, and Li}]{wang-etal-2024-leave}
Minzheng Wang, Longze Chen, Fu~Cheng, Shengyi Liao, Xinghua Zhang, Bingli Wu, Haiyang Yu, Nan Xu, Lei Zhang, Run Luo, Yunshui Li, Min Yang, Fei Huang, and Yongbin Li. 2024.
\newblock \href {https://doi.org/10.18653/v1/2024.emnlp-main.322} {Leave no document behind: Benchmarking long-context {LLM}s with extended multi-doc {QA}}.
\newblock In \emph{Proceedings of the Conference on Empirical Methods in Natural Language Processing}.

\bibitem[{Wei et~al.(2024)Wei, Karina, Chung, Jiao, Papay, Glaese, Schulman, and Fedus}]{Wei2024MeasuringSF}
Jason Wei, Nguyen Karina, Hyung~Won Chung, Yunxin~Joy Jiao, Spencer Papay, Amelia Glaese, John Schulman, and William Fedus. 2024.
\newblock \href {https://arxiv.org/abs/2411.04368} {Measuring short-form factuality in large language models}.
\newblock \emph{Preprint}, arXiv:2411.04368.

\bibitem[{Ye et~al.(2024)Ye, Sun, Arik, and Pfister}]{ye-etal-2024-effective}
Xi~Ye, Ruoxi Sun, Sercan Arik, and Tomas Pfister. 2024.
\newblock \href {https://aclanthology.org/2024.naacl-long.346.pdf} {Effective large language model adaptation for improved grounding and citation generation}.
\newblock In \emph{Proceedings of the Conference of the North American Chapter of the Association for Computational Linguistics}.

\bibitem[{Ye et~al.(2025)Ye, Yin, He, Zhang, Yen, Gao, Durrett, and Chen}]{Ye-Et-Al:2025:Longproc}
Xi~Ye, Fangcong Yin, Yinghui He, Joie Zhang, Howard Yen, Tianyu Gao, Greg Durrett, and Danqi Chen. 2025.
\newblock \href {https://arxiv.org/abs/2501.05414} {Longproc: Benchmarking long-context language models on long procedural generation}.
\newblock \emph{Preprint}, arXiv:2501.05414.

\bibitem[{Yona et~al.(2024)Yona, Aharoni, and Geva}]{Yona2024NarrowingTK}
Gal Yona, Roee Aharoni, and Mor Geva. 2024.
\newblock \href {https://doi.org/10.18653/v1/2024.acl-long.365} {Narrowing the knowledge evaluation gap: Open-domain question answering with multi-granularity answers}.
\newblock In \emph{Proceedings of the Annual Meeting of the Association for Computational Linguistics}.

\bibitem[{Yoran et~al.(2024)Yoran, Wolfson, Ram, and Berant}]{yoran2024making}
Ori Yoran, Tomer Wolfson, Ori Ram, and Jonathan Berant. 2024.
\newblock \href {https://openreview.net/forum?id=ZS4m74kZpH} {Making retrieval-augmented language models robust to irrelevant context}.
\newblock In \emph{Proceedings of the International Conference on Learning Representations}.

\bibitem[{Zeng and Gao(2023)}]{Zeng2023PromptTB}
Fengzhu Zeng and Wei Gao. 2023.
\newblock \href {https://api.semanticscholar.org/CorpusID:259076388} {Prompt to be consistent is better than self-consistent? few-shot and zero-shot fact verification with pre-trained language models}.
\newblock In \emph{Proceedings of the Association for Computational Linguistics}.

\bibitem[{Zhang et~al.(2025)Zhang, Li, Zeng, and Wang}]{Zhang2024JasperAS}
Dun Zhang, Jiacheng Li, Ziyang Zeng, and Fulong Wang. 2025.
\newblock \href {https://arxiv.org/abs/2412.19048} {Jasper and stella: distillation of sota embedding models}.
\newblock \emph{Preprint}, arXiv:2412.19048.

\bibitem[{Zhang and Gao(2023)}]{Zhang2023TowardsLF}
Xuan Zhang and Wei Gao. 2023.
\newblock \href {https://api.semanticscholar.org/CorpusID:263334529} {Towards llm-based fact verification on news claims with a hierarchical step-by-step prompting method}.
\newblock In \emph{Proceedings of the International Joint Conference on Natural Language Processing}.

\bibitem[{Zhang et~al.(2023)Zhang, Li, Cui, Cai, Liu, Fu, Huang, Zhao, Zhang, Chen, Wang, Luu, Bi, Shi, and Shi}]{Zhang2023SirensSI}
Yue Zhang, Yafu Li, Leyang Cui, Deng Cai, Lemao Liu, Tingchen Fu, Xinting Huang, Enbo Zhao, Yu~Zhang, Yulong Chen, Longyue Wang, Anh~Tuan Luu, Wei Bi, Freda Shi, and Shuming Shi. 2023.
\newblock \href {https://arxiv.org/abs/2309.01219} {Siren's song in the ai ocean: A survey on hallucination in large language models}.
\newblock \emph{Preprint}, arXiv:2309.01219.

\bibitem[{Zheng et~al.(2023)Zheng, Chiang, Sheng, Zhuang, Wu, Zhuang, Lin, Li, Li, Xing, Zhang, Gonzalez, and Stoica}]{zheng2023judging}
Lianmin Zheng, Wei-Lin Chiang, Ying Sheng, Siyuan Zhuang, Zhanghao Wu, Yonghao Zhuang, Zi~Lin, Zhuohan Li, Dacheng Li, Eric.~P Xing, Hao Zhang, Joseph~E. Gonzalez, and Ion Stoica. 2023.
\newblock \href {https://arxiv.org/abs/2306.05685} {Judging llm-as-a-judge with mt-bench and chatbot arena}.
\newblock \emph{Preprint}, arXiv:2306.05685.

\bibitem[{Zhong et~al.(2022)Zhong, Shi, tau Yih, and Zettlemoyer}]{Zhong2022RoMQAAB}
Victor Zhong, Weijia Shi, Wen tau Yih, and Luke Zettlemoyer. 2022.
\newblock \href {https://api.semanticscholar.org/CorpusID:253116788} {Romqa: A benchmark for robust, multi-evidence, multi-answer question answering}.
\newblock In \emph{Proceedings of the Conference on Empirical Methods in Natural Language Processing}.

\bibitem[{Zhou et~al.(2023)Zhou, Sch{\"a}rli, Hou, Wei, Scales, Wang, Schuurmans, Cui, Bousquet, Le, and Chi}]{zhou2023leasttomost}
Denny Zhou, Nathanael Sch{\"a}rli, Le~Hou, Jason Wei, Nathan Scales, Xuezhi Wang, Dale Schuurmans, Claire Cui, Olivier Bousquet, Quoc~V Le, and Ed~H. Chi. 2023.
\newblock \href {https://openreview.net/forum?id=WZH7099tgfM} {Least-to-most prompting enables complex reasoning in large language models}.
\newblock In \emph{Proceedings of the International Conference on Learning Representations}.

\end{thebibliography}
